\documentclass[10pt,journal,compsoc]{IEEEtran}
\ifCLASSOPTIONcompsoc
  \usepackage[nocompress]{cite}
\else
  \usepackage{cite}
\fi
\ifCLASSINFOpdf
  \usepackage[pdftex]{graphicx}
  \graphicspath{{pdf/}{png/}{jpg/}}
\else
  \usepackage[dvips]{graphicx}
  \graphicspath{{../eps/}}
  \DeclareGraphicsExtensions{.eps}
\fi
\usepackage{amsmath}
\usepackage{array}

\ifCLASSOPTIONcompsoc
  \usepackage[caption=false,font=footnotesize,labelfont=sf,textfont=sf]{subfig}
\else
  \usepackage[caption=false,font=footnotesize]{subfig}
\fi

\usepackage{stfloats}
\fnbelowfloat
\ifCLASSOPTIONcaptionsoff
  \usepackage[nomarkers]{endfloat}
 \let\MYoriglatexcaption\caption
 \renewcommand{\caption}[2][\relax]{\MYoriglatexcaption[#2]{#2}}
\fi
\usepackage{url}

\usepackage{macro}
\usepackage{amsfonts}
\usepackage{multirow}

\begin{document}
%
\title{Deep Non-Rigid Structure from Motion \\ with Missing Data}
%
%
%
%

\author{Chen~Kong,~\IEEEmembership{Student Member,~IEEE,}
        and~Simon~Lucey,~\IEEEmembership{Member,~IEEE}
\IEEEcompsocitemizethanks{\IEEEcompsocthanksitem C. Kong and S. Lucey are with
the Robotics Institute, Carnegie Mellon University, 5000 Forbes Ave, Pittsburgh,
PA, 15213.\protect\\
E-mail: \{chenk, slucey\}@cs.cmu.edu}
\thanks{Manuscript received April 19, 2005; revised August 26, 2015.}}

%
%

\markboth{IEEE TRANSACTIONS ON PATTERN ANALYSIS AND MACHINE INTELLIGENCE,~Vol.~14, No.~8, August~2015}%
{Shell \MakeLowercase{\textit{et al.}}: Bare Demo of IEEEtran.cls for Computer Society Journals}
%



\IEEEtitleabstractindextext{%
\begin{abstract}
Non-Rigid Structure from Motion (NRSfM) refers to the problem of reconstructing
cameras and the 3D point cloud of a non-rigid object from an ensemble of images
with 2D correspondences. Current NRSfM algorithms are limited from two perspectives: (i) the number of images, and (ii) the type of shape variability they can handle. These difficulties stem from the inherent conflict between the condition of the system and the degrees of freedom needing to be modeled -- which has hampered its practical utility for many applications within vision. In this paper we propose a novel hierarchical sparse coding model for NRSFM which can overcome (i) and (ii) to such an extent, that NRSFM can be applied to problems in vision previously thought too ill posed. Our approach is realized in practice as the training of an unsupervised deep neural network (DNN) auto-encoder with a unique architecture that is able to disentangle pose from 3D structure. Using modern deep learning computational platforms allows us to solve NRSfM problems at an unprecedented scale and shape complexity. Our approach has no 3D supervision, relying solely on 2D point correspondences. Further, our approach is also able to handle missing/occluded 2D points without the need for matrix completion. Extensive experiments demonstrate the impressive performance of our approach where we exhibit superior precision and robustness against all available state-of-the-art works in some instances by an order of magnitude.  We further propose a new quality measure (based on the network weights) which circumvents the need for 3D ground-truth to ascertain the confidence we have in the reconstructability. We believe our work to be a significant advance over state-of-the-art in NRSFM. 
\end{abstract}

\begin{IEEEkeywords}
Nonrigid structure from motion, hierarchical sparse coding, deep neural network, reconstructability, 
missing data.
\end{IEEEkeywords}}

\maketitle

\IEEEdisplaynontitleabstractindextext

%
\IEEEpeerreviewmaketitle

\IEEEraisesectionheading{\section{Introduction}\label{sec:introduction}}

%
%
%
%
\IEEEPARstart{B}{uilding} an AI capable of inferring the 3D structure and pose of an object from
a single image is a problem of immense importance. Training such a system using
supervised learning requires a large number of labeled images~--~how to obtain
these labels is currently an open problem for the vision community.
Rendering~\cite{su2015render} is problematic as the synthetic images seldom
match the appearance and geometry of the objects we encounter in the
real-world. Hand annotation is preferable, but current strategies rely on
associating the natural images with an external 3D dataset 
(\eg~ShapeNet~\cite{DBLP:journals/corr/ChangFGHHLSSSSX15}, ModelNet~\cite{wu20153d}),
which we refer to as \emph{3D supervision}. If the 3D shape dataset does not
capture the variation we see in the imagery, then the problem is inherently
ill-posed.

Non-Rigid Structure from Motion (\nrsfm{}) offers computer vision a way out of
this quandary -- by recovering the pose and 3D structure of an object category
\emph{solely} from hand annotated 2D landmarks with no need of 3D supervision.
Classically~\cite{bregler2000recovering}, the problem of \nrsfm{} has been applied
to objects that move non-rigidly over time such as the human body and face. Additional benchmarks have been proposed~\cite{jensen2018benchmark} for other temporally deforming non-rigid objects. But \nrsfm{} is not restricted to non-rigid objects; it can equally be applied to rigid objects whose object categories are non-rigid~\cite{kong2016sfc, agudo2018image, vicente2014reconstructing}. Consider, for example, the five objects in \figurename~\ref{fig:teaser}~(top), instances from
the visual object category ``chair''. Each object in
isolation represents a rigid chair, but the set of all 3D shapes describing
``chair'' is non-rigid. In other words, each object instance can be modeled as
a certain deformation from its category's general shape.

Rigid \sfm{} is already an ill-posed problem. It is the rigidity prior of objects that helps
to obtain good results across \sfm{} applications. \nrsfm{} is even more challenging where rigidity
is removed due to its inherent non-rigid nature.
To resolve the problem of \nrsfm{}, additional shape priors are proposed, \eg~low
rank~\cite{bregler2000recovering, dai2014simple},
union-of-subspaces~\cite{zhu2014complex, agudo2018image},
and block-sparsity~\cite{kong2016prior, kong2016sfc}. However, low rank is only applicable to
simple non-rigid objects with limited deformations and union-of-subspaces rely heavily on frame
clustering which has difficulty scaling up to large image collections. A block-sparsity prior where each shape can be represented by at most $K$ bases out of $L$, is considered as one of 
the most promising assumptions in terms of
covering broad shape variations. This is because sparsity can be thought as a union of
$\binom{L}{K}$ subspaces where $L$ could be large then an over-complete dictionary is utilized.
However, pointed by our previous work~\cite{kong2016prior}, searching the best subspace out of
$\binom{L}{K}$ is extremely hard and not robust. Based on this observation, in this paper,
we propose a novel shape prior using hierarchical sparse coding. The introduced additional
layers compared to single-layer sparse coding are capable of controlling the number of subspaces
by learning from data such that invalid subspaces are removed while sufficient subspaces are
remained for modeling shape variations. This insight is at the heart of our paper.



\subsection{Contributions}
We propose a novel shape prior based on hierarchical sparse coding and demonstrate that the 
2D projections under weak perspective cameras can be represented by the hierarchical 
dictionaries in a block sparse way. Through recent theoretical innovations~\cite{papyan2017convolutional}, we then show how this problem can be reinterpreted as training an unsupervised feed-forward Deep Neural Network(DNN) auto-encoder.
A common drawback of DNNs when applied to reconstruction problems is that they are an opaque black-box lacking any interpretability. A strength of our approach is that the DNN is directly derived from a hierarchical block sparse dictionary learning objective -- allowing for greater transparency into what the network weights are modeling. As a result we are able to formulate a measure of model quality (using the coherence of learned parameters), which helps to avoid over-fitting especially when ground-truth of training data are not available.

\begin{figure}[!t]
\centering
\includegraphics[width=\linewidth]{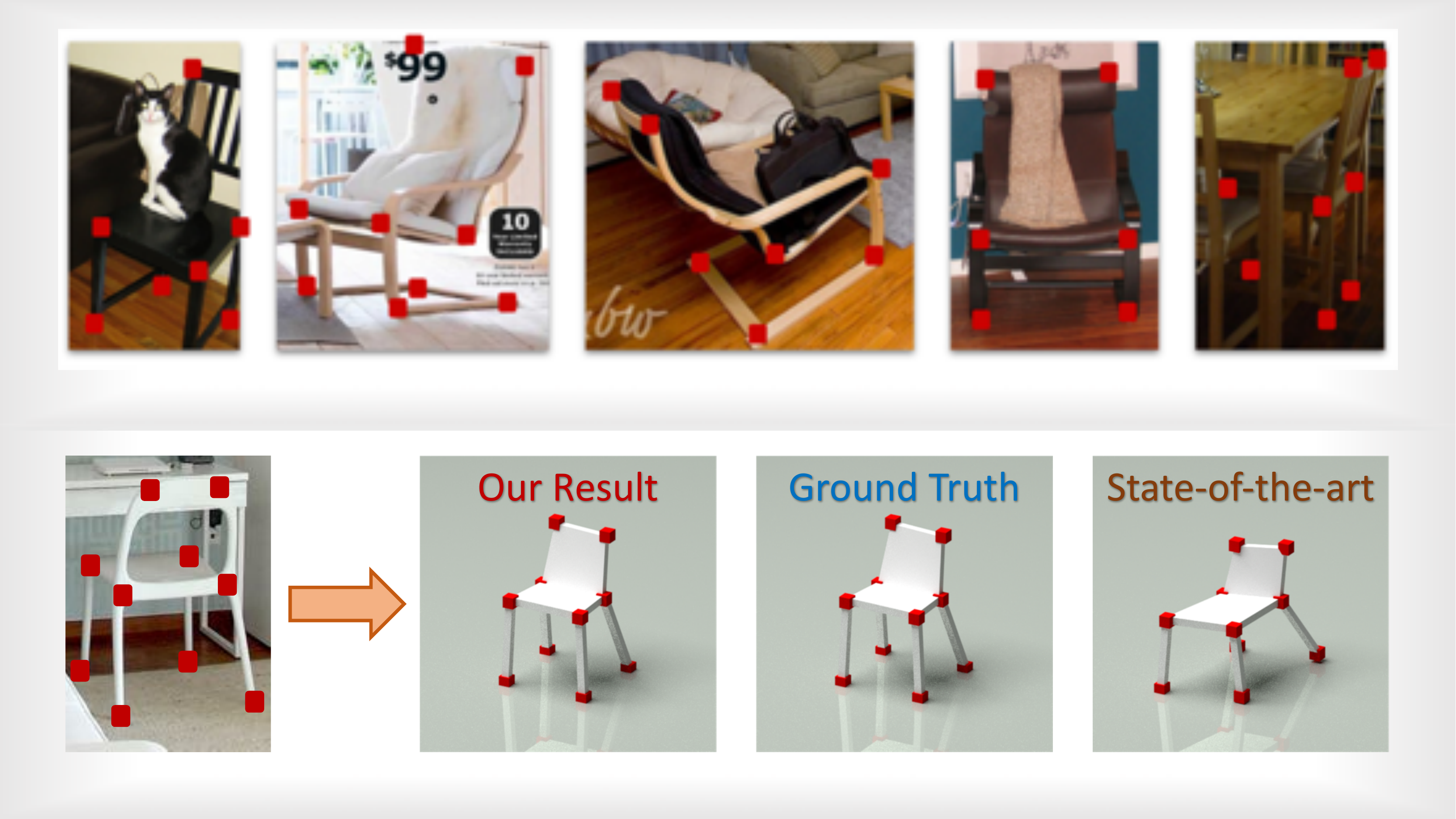}
\caption{In this paper, we want to reconstruct 3D shapes solely from a sequence of 
annotated images---shown on the top---with no need of 3D ground truth. Our proposed
hierarchical sparse coding model and corresponding deep solution outperform state-of-the-arts
in the order of magnitude.}
\label{fig:teaser}
\end{figure}

Our deep \nrsfm{} is capable of handling hundreds of thousands of images and learning large
parameterizations to model non-rigidity.  Extensive experiments are conducted and our approach outperforms state-of-the-art methods in the order of magnitude on a number of benchmarks. Both quantitative and qualitative results demonstrate our superior performance -- an example of qualitative results is shown in \figurename~\ref{fig:teaser}.  

Compared to our previous work~\cite{kong2019deep}, this paper offers a substantial leap forward in terms of state-of-the-art. Our approach can now handle ``real-world'' scenarios where camera scale and translation are unknown and there are missing or occluded 2D points. Specifically, the method is now capable of reconstructing non-rigid objects under weak perspective projection instead of orthogonal (as in~\cite{kong2019deep}). Weak perspective projection is a reasonable assumption for many practical vision applications where the object's variation in depth is small compared to their distance from the 
camera~\cite{tomasi1992shape, bregler2000recovering,xiao2006closed, xiang2014beyond}. 
Additionally, unlike our earlier work~\cite{kong2019deep}, this paper provides a solution to 
reconstruct invisible points -- 2D coordinates missing due to occlusion or self-occlusion -- with no need of matrix completion. These two breakthroughs make this paper substantially different from the deep \nrsfm{} first proposed in~\cite{kong2019deep} and dramatically improve its practical utilities in real-world applications.

\subsection{Related Work}
In rigid structure from motion, the rank of 3D structure is fixed,~\cite{tomasi1992shape}
since the 3D structure remains the same between frames. Based on this insight, Bregler~\etal~\cite{bregler2000recovering} advocated that non-rigid 3D structure could be represented by a linear subspace of low rank. Dai~\etal~\cite{dai2014simple} developed this prior further by proving that the low-rank assumption itself is sufficient to address the ill-posedness of
\nrsfm{} with no need of additional priors. They then proposed a novel algorithm based on singular value decomposition, followed by a non-linear optimization to achieve state-of-the-art performance across numerous benchmarks. The low-rank assumption has also been applied temporally~\cite{akhter2009defense, fragkiadaki2014grouping} -- 3D point trajectories can be represented by pre-defined (\eg~DCT) or learned bases. Although exhibiting impressive performance, the low-rank assumption has a major drawback. The rank is strictly 
limited by the number of points and frames (whichever is smaller~\cite{dai2014simple}). This makes low-rank \nrsfm~infeasible if we want to solve large-scale problems with -- complex shape variations -- when the number of points is substantially smaller than the number of frames. 

Inspired by the intuition that complex non-rigid deformations could be clustered into
a sequence of simple motions, Zhu~\etal~\cite{zhu2014complex} proposed to model non-rigid
3D structure by a union of local subspaces.  They show that clustering frames from their 2D
annotations is less effective and therefore propose a novel algorithm to reconstruct 3D shapes and an estimate of the 3D-based frame affinity matrix simultaneously.
The idea of union-of-subspaces was later extended to spatial-temporal domain~\cite{agudo2017dust} and applied to rigid object category reconstruction~\cite{agudo2018image}.Though the union-of-subspaces model is capable of handling complex object deformation, its holistic
estimation of the entire affinity matrix -- the number of frame by the number of frame matrix -- impedes its scalability to large-scale problems \eg~more than tens of thousand frames. 

Inspired by the union-of-subspaces assumption, a block-sparsity prior~\cite{kong2016prior, zhou2015sparseness, kong2016sfc} was proposed as a more generic and elegant prior for \nrsfm{}. The sparsity prior can be viewed as being mathematically equivalent to the union 
of \emph{all} possible local subspaces. This is advantageous as it circumvents the need for a messy affinity matrix -- since all local subspaces are being considered. Further, since the method is entertaining many local subspaces it is also able to model much more complex 3D shapes than single subspace low-rank methods. However, the sheer number of subspaces that can be entertained by the block-sparsity prior is its fundamental drawback. Since there are so many possible subspaces to choose from, the approach is quite sensitive to noise dramatically limiting its applicability to ``real-world'' \nrsfm{} problems. In this paper we want to leverage the elegance and expressibility of the block-sparsity prior without suffering from its inherent sensitivity to noise. 

Drover~\etal~\cite{drover2018can} recently applied neural networks for \nrsfm{} in an unsupervised manner.  Drover~\etal{} used a GAN to generate potential 3D structures and ensure their consistency across novel viewpoints through the GAN's discriminator.  Although intriguing, this approach to date has only been shown effective for 3D human skeletons --  whereas our approach exhibits state-of-the-art performance across multiple disparate object classes.  Further, unlike~\cite{drover2018can}, the connection of our approach to sparse dictionary learning  facilitates  a  measure  (based  on  dictionary  coherence) of the uniqueness of the 3D reconstruction without having any 3D ground-truth.  Finally, it is unclear how Drover~\etal{} handles missing points -- again limiting the applicability of their approach to to ``real-world'' problems. 

Missing correspondences are inevitable when annotating objects because key points are often occluded by other objects or itself, shown in \figurename~\ref{fig:teaser}~(top). This requires any practical \nrsfm{} algorithm to be tolerant to missing 2D points. Two solutions are usually utilized: (i) a mask is introduced indicating
the visibility of 2D points to the objective function~\cite{gotardo2011kernel, del2007non, kong2016sfc}, or (ii) the employment of a matrix completion algorithm to recover the missing 2D points a priori from which then the \nrsfm{} algorithm~\cite{dai2014simple, lee2016consensus, hamsici2012learning} is applied. In this paper, we follow the
former strategy and show how this is implemented by a feed-forward neural network.

\subsection{Background}
\label{sec: scdnn}
Sparse dictionary learning can be considered as an unsupervised learning task
and divided into two sub-problems: (i) dictionary learning, and (ii) sparse code recovery. 
Let us consider sparse code recovery problem, where we estimate a sparse representation $\zv$ for 
a measurement vector $\xv$ given the dictionary~$\Dv$, \ie
\begin{equation}
 \min_\zv \Vert \xv - \Dv\zv\Vert_2^2 \quad \st \Vert \zv \Vert_0 < \lambda,
 \label{eq:sparse_coding}
\end{equation}
where $\lambda$ related to the trust region controls the sparsity of recovered
code. One classical algorithm to recover the sparse representation is
Iterative Shrinkage and Thresholding Algorithm
(ISTA)~\cite{daubechies2004iterative, rozell2008sparse,beck2009fast}. ISTA
iteratively executes the following two steps with $\zv^{[0]} = \zero$:
\begin{gather}
 \vv = \zv^{[i]} - \alpha\Dv^T(\Dv\zv^{[i]} - \xv), \\
 \zv^{[i+1]} = \argmin_{\uv} \frac{1}{2}\Vert \uv - \vv \Vert^2_2 + \tau\Vert\uv\Vert_1,
\end{gather}
which first uses the gradient of~$\Vert \xv - \Dv\zv\Vert_2^2$ to
update~$\zv^{[i]}$ in step size $\alpha$ and then finds the closest sparse
solution using an $\ell_1$ convex relaxation. It can be demonstrated that the second step has 
a closed-form solution that is
\begin{equation}
    \zv^{[i+1]} = \eta(\vv; \tau).
    \label{eq:st}
\end{equation}
where $\eta$ represents a element-wise soft-thresholding operation, formally defined as
\begin{equation}
    \eta(x; b) = \begin{cases}
        x - b & \text{if}~x>b, \\
        x + b & \text{if}~x<-b, \\
        0 & \text{otherwise}.
    \end{cases}
    \label{eq:eta}
\end{equation}
Therefore, ISTA can be summarized as the following recursive equation:
\begin{equation}
 \zv^{[i+1]} = \eta\big(\zv^{[i]} - \alpha\Dv^T(\Dv\zv^{[i]} - \xv); \tau\big),
 \label{eq:ista}
\end{equation}
where $\tau$ is related to $\lambda$ for controlling sparsity.

Recently, Papyan~\cite{papyan2017convolutional} proposed to use ISTA and sparse
coding to reinterpret feed-forward neural networks. They argue that feed-forward
passing a single-layer neural network~$\zv = \eta(\Dv^T\xv; b)$ can be
considered as one iteration of ISTA in (\ref{eq:ista}) when setting~$\alpha=1$ and~$\tau = b$.
Based on this insight, the authors extend this interpretation to feed-forward
neural network with~$N$ layers
\begin{equation}
 \begin{aligned}
  \zv_1 & = \eta(\Dv_1^T\xv; b_1)       \\
  \zv_2 & = \eta(\Dv_2^T\zv_1; b_2)     \\
        & \quad \vdots                    \\
  \zv_N & = \eta(\Dv_N^T\zv_{N-1}; b_N) \\
 \end{aligned}
\end{equation}
as executing a sequence of single-iteration ISTA, serving as an approximate
solution to the hierarchical sparse coding problem: find~$\{\zv_i\}_{i=1}^N$,
such that
\begin{equation}
 \begin{aligned}
  \xv = \Dv_1\zv_1       & , \quad \Vert \zv_1 \Vert_0 < \lambda_1, \\
  \zv_1 = \Dv_2\zv_2     & , \quad \Vert \zv_2 \Vert_0 < \lambda_2, \\
  \vdots \quad \quad     & , \quad \quad \vdots                                  \\
  \zv_{N-1} = \Dv_N\zv_N & , \quad \Vert \zv_N \Vert_0 < \lambda_N, \\
 \end{aligned}
\end{equation}
where the bias terms~$\{b_i\}_{i=1}^N$ (in a similar manner to $\tau$) are
related to~$\{\lambda_i\}_{i=1}^N$, adjusting the sparsity of recovered code.
Furthermore, they reinterpret back-propagating through the deep neural network
as learning the dictionaries~$\{\Dv_i\}_{i=1}^N$. This connection offers a novel reinterpretation of DNNs through the lens of hierarchical sparse dictionary learning. In this paper, we extend this reinterpretation to the block sparse scenario and apply it to solving our \nrsfm{} problem.

\section{Problem Formulation}
In the context of \nrsfm{}, the weak perspective projection model is a reasonable assumption since
the many of objects we deal with in vision applications 
have a much smaller depth variation compared to their distance from the camera. We shall start from the orthogonal projection model in this section and
then generalize to weak perspective projection in Section~\ref{sec:weak-perspective}.
Under orthogonal projection, \nrsfm{} deals with the problem of factorizing
a 2D projection matrix $\Wv\in\RR^{P\times 2}$, given $P$ points, as the product of a 3D shape
matrix $\Sv\in\RR^{P\times 3}$ and a camera matrix $\Mv\in\RR^{3\times 2}$.
Formally,
\begin{equation}
 \Wv = \Sv\Mv,
 \label{eq:proj}
\end{equation}
\begin{equation}
 \Wv = \begin{bmatrix} u_1 & v_1 \\ u_2 & v_2 \\ \vdots & \vdots \\ u_P & v_P \end{bmatrix},~
 \Sv = \begin{bmatrix} x_1 & y_1 & z_1 \\ x_2 & y_2 & z_2 \\ \vdots & \vdots & \vdots \\ x_P & y_P & z_P \end{bmatrix},~
 \Mv^T\Mv = \Iv_2,
\end{equation}
where $(u_i, v_i)$ and $(x_i, y_i, z_i)$ are the image and world coordinates of the
$i$-th point respectively. The goal of \nrsfm{} is to recover simultaneously
the shape~$\Sv$ and the camera $\Mv$ for each projection $\Wv$ in a given set
$\mathbb{W}$ of 2D landmarks. In a general \nrsfm{} including \sfc{}, this
set~$\mathbb{W}$ could contain deformations of a non-rigid object or various
instances from an object category.

\section{Modeling via hierarchical sparse coding}
\label{sec:mlscm}
Kong et al.~\cite{kong2016prior} argued that an effective solution for \nrsfm{} can be found by assuming the vectorization of $\Sv$ can be represented
by a dictionary sparsely:
\begin{equation}
\label{eq:single-layer-sparse-coding}
\sv = \Dv\psiv, \quad \Vert \psiv \Vert_0 < \lambda \;\;. \end{equation}
This paper introduces additional layers and therefore a hierarchical sparse model is proposed:
\begin{equation}
 \begin{aligned}
  \sv = \Dv_1\psiv_1         & , \quad \Vert \psiv_1 \Vert_0 < \lambda_1, \\
  \psiv_1 = \Dv_2\psiv_2     & , \quad \Vert \psiv_2 \Vert_0 < \lambda_2, \\
  \vdots \quad \quad         & , \quad \quad \vdots                       \\
  \psiv_{N-1} = \Dv_N\psiv_N & , \quad \Vert \psiv_N \Vert_0 < \lambda_N, \\
 \end{aligned}
 \label{eq:mlsc}
\end{equation}
where $\Dv_1 \in \RR^{3P\times K_1}, \Dv_2 \in \RR^{K_1 \times K_2}, \dots,
 \Dv_N \in \RR^{K_{N-1}\times K_N}$ are hierarchical dictionaries and $\psiv_1\in\RR^{K_1}$,
$\psiv_2\in\RR^{K_2}$, $\dots$, $\psiv_N\in\RR^{K_N}$ are hierarchical sparse codes.
In this prior, each non-rigid shape is represented by a sequence of dictionaries and corresponding
non-negative sparse codes hierarchically. Each sparse code is determined by its lower-level 
neighbor and affects the next-level. The additional layers introduced by this hierarchy
increase the number of variables, and thus increase the degree of freedom of the system. However, 
these additional layers actually result in a more constrained and thus stable sparse code recovery
process.

Sparse code recovery algorithms in general attempt to solve two problems: 1) select the best
subspace and 2) estimate the closest representation within the subspace. These two problems could be
solved simultaneously or alternatively, but the quality of recovered sparse code highly relies on the 
former. If the desired subspace is given from oracle, then the sparse coding problem degenerates
to a linear system. However, without knowing the 
size of the desired subspace, the number of valid subspaces in (\ref{eq:single-layer-sparse-coding}) is combinatorial to the number of dictionary atoms $K$
\ie~$\sum_{n=1}^K\binom{K}{n}$. Selecting the best subspace out of such large number of candidates is
considerably difficult, especially when using over-complete dictionaries. This reveals the 
conflict between the quality of sparse code recovery and the representing capacity of the dictionary, 
and further explains the sensitivity of~\cite{kong2016prior} to non-compressible sequences.

The additional layers introduced in this paper alleviate the dilemma. In (\ref{eq:mlsc}), the
sparse code $\psiv_1$ is not completely free but represented by the subsequent dictionaries. Therefore,
the number of subspaces is not combinatorial to $K_1$ but controlled by the subsequent dictionaries
$\{\Dv_i\}_{i=2}^N$. If the subsequent dictionaries are learned properly, they could serve as a filter
so that only functional subspaces remain and redundant ones are removed. This directly breaks the
combinatorial explosion of the number of subspaces and consequently maintains the robustness of sparse
code recovery. Based on this observation, we are able to utilize substantially over-complete dictionaries
to model a highly deformable object from a large scale image collection with no worries about
reconstructability and robustness.

\subsection{Hierarchical Block Sparse Coding}
\label{sec:hbsc}
Given the proposed hierarchical sparse coding model, shown in (\ref{eq:mlsc}), we now build a 
conduit from the 2D correspondences $\Wv$ to the proposed shape code~$\{\psiv_i\}_{i=1}^k$.
Since $\sv\in\RR^{3P}$ in (\ref{eq:mlsc}) is the vectorization of $\Sv\in\RR^{P\times 3}$, it can be well modeled via  \ie~$\Sv = \Dv^\sharp_1(\psiv_1 \otimes \Iv_3)$ where $\otimes$ is the Kronecker product and
$\Dv_1^\sharp\in\RR^{P\times 3K_1}$ is a reshape of $\Dv_1\in\RR^{3P\times K_1}$~\cite{dai2014simple}.
It is known that~$\Av\Bv \otimes \Iv = (\Av\otimes \Iv)(\Bv\otimes\Iv)$
given two matrices~$\Av, \Bv$, and identity matrix $\Iv$. Based on this lemma, we can derive
that
\begin{equation}
 \begin{aligned}
  \Sv = \Dv^\sharp_1(\psiv_1 \otimes \Iv_3)                               & , \quad \Vert \psiv_1 \Vert_0 < \lambda_1, \\
  \psiv_1\otimes \Iv_3 = (\Dv_2 \otimes \Iv_3)(\psiv_2\otimes \Iv_3)      & , \quad \Vert \psiv_2 \Vert_0 < \lambda_2,\\
  \vdots \quad \quad                                                      & , \quad \quad \vdots                                      \\
  \psiv_{N-1} \otimes \Iv_3 = (\Dv_N \otimes \Iv_3)(\psiv_N\otimes \Iv_3) & , \quad \Vert \psiv_N \Vert_0 < \lambda_N. \\
 \end{aligned}
 \label{eq:mlbsc_s}
\end{equation}

Further, from (\ref{eq:proj}), by right multiplying the camera matrix~$\Mv\in\RR^{3\times2}$ to the 
both sides of (\ref{eq:mlbsc_s}) and denote $\Psiv_i = \psiv_i \otimes \Mv$, we obtain that
\begin{equation}
 \small
 \begin{aligned}
  \Wv = \Dv^\sharp_1 \Psiv_1                 & , \quad \Vert \Psiv_1 \Vert_0^{(3\times 2)} < \lambda_1, \\
  \Psiv_1 = (\Dv_2 \otimes \Iv_3)\Psiv_2     & , \quad \Vert \Psiv_2 \Vert_0^{(3\times 2)} < \lambda_2, \\
  \vdots \quad \quad                         & , \quad \quad \vdots                                     \\
  \Psiv_{N-1} = (\Dv_N \otimes \Iv_3)\Psiv_N & , \quad \Vert \Psiv_N \Vert_0^{(3\times 2)} < \lambda_N, \\
 \end{aligned}
 \label{eq:mlbsc_w}
\end{equation}
where $\Vert \cdot \Vert_0^{(3\times 2)}$ divides the argument matrix into
blocks with size $3\times 2$ and counts the number of active blocks.
Since~$\psiv_i$ has active elements less than $\lambda_i$, $\Psiv_i$ has active
blocks less than $\lambda_i$, that is $\Psiv_i$ is block sparse.
This derivation demonstrates that if the shape vector $\sv$ satisfies the
hierarchical sparse coding prior described by (\ref{eq:mlsc}), then its
2D projection~$\Wv$ must be in the format of hierarchical \emph{block} sparse
coding described by (\ref{eq:mlbsc_w}). We hereby interpret \nrsfm{}
as a hierarchical \emph{block} sparse dictionary learning problem, \ie~factorizing $\Wv$ as products of hierarchical dictionaries~$\{\Dv_i\}_{i=1}^N$ and block sparse coefficients~$\{\Psiv_i\}_{i=1}^N$.

\section{Deep Neural Network Solution}
\label{sec:architecture}

\begin{figure*}[!t]
\centering
\includegraphics[width=\linewidth]{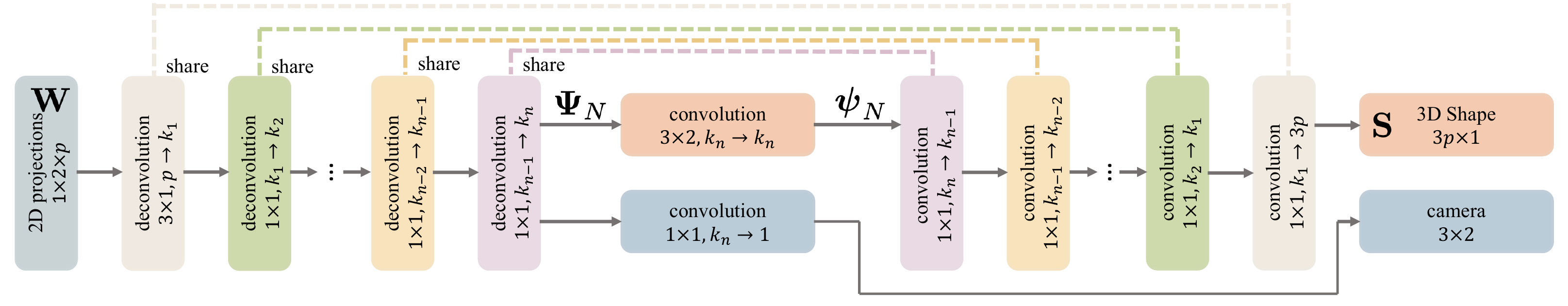}
\caption{Architecture of our proposed deep \nrsfm{}. The network can be divided into
1) Encoder: from 2D correspondences $\Wv$ to the hidden block sparse code $\Psiv_N$, 
2) Bottleneck: from hidden block sparse code $\Psiv_N$ to hidden regular sparse code $\psiv_N$ and camera,
3) Decoder: from hidden regular sparse code $\psiv_N$ to 3D reconstructed shape $\Sv$.
The encoder and decoder are intentionally designed to share convolution kernels (\ie~dictionaries) and
form a symmetric formulation. The symbol $a\times b, c \to d$ refers to the convolution layer using
kernel size $a \times b$ with $c$ input channels and $d$ output channels.}
\label{fig:architecture}
\end{figure*}

Before solving the hierarchical block sparse coding problem in (\ref{eq:mlbsc_w}),
we first consider a single-layer problem:
\begin{equation}
 \min_{\Zv} \Vert \Xv - \Dv\Zv\Vert_F^2 \quad \st~\Vert \Zv \Vert_{0}^{(3\times2)} < \lambda.
\end{equation}
Inspired by ISTA, we propose to solve this problem by iteratively executing
the following two steps:
\begin{gather}
 \Vv = \Zv^{[i]} - \alpha\Dv^T(\Dv\Zv^{[i]} - \Xv), \\
 \Zv^{[i+1]} = \argmin_{\Uv} \frac{1}{2}\Vert \Uv - \Vv \Vert^2_F + \tau\Vert\Uv\Vert_{F1}^{(3\times2)},
 \label{eq:block-st}
\end{gather}
where $\Vert \cdot \Vert_{F1}^{(3\times2)}$ is defined as the summation of the
Frobenius norm of each $3\times2$ block, serving as a convex relaxation of the
block sparsity constraint. 
Recall the regular sparse situation in Section~\ref{sec: scdnn}.
Analogous to (\ref{eq:st}), we use an approximate solution to (\ref{eq:block-st}) for computational
efficiency, \ie
\begin{equation}
\Zv^{[i+1]} = \eta(\Vv; \bv\otimes\one_{3\times2}),
\label{eq:block-relu}
\end{equation}
where $\eta$ represents a element-wise soft-thresholding operation defined in (\ref{eq:eta}), 
$\one_{3\times2}$ denotes a 3-by-2 matrix filled with one and $\bv$ is a vector that controls the 
trust region for each block.
Based on this approximation, a single-iteration block ISTA with step size
$\alpha=1$ can be represented by :
\begin{equation}
 \Zv = \eta(\Dv^T\Xv; \bv\otimes\one_{3\times 2}),
 \label{eq:singleBISTA}
\end{equation}

\subsection{Encoder}
Recall from Section~\ref{sec: scdnn} that the feed-forward pass through a deep
neural network can be considered as a sequence of single ISTA iterations and
thus provides an approximate recovery of hierarchical sparse codes. We follow
the same scheme -- sequentially using single-iteration block ISTA -- to solve 
the hierarchical block sparse coding problem (\ref{eq:mlbsc_w}) \ie
\begin{equation}
 \begin{aligned}
  \Psiv_1 & = \eta((\Dv^\sharp_1)^T\Wv; \bv_1\otimes\one_{3\times 2}),                \\
  \Psiv_2 & = \eta((\Dv_2 \otimes \Iv_3)^T\Psiv_1; \bv_2\otimes\one_{3\times 2}),     \\
          & \quad \vdots                                                                \\
  \Psiv_N & = \eta((\Dv_N \otimes \Iv_3)^T\Psiv_{N-1}; \bv_N\otimes\one_{3\times 2}), \\
 \end{aligned}
\end{equation}
where $\{\bv_i\}_{i=1}^N$ are learnable parameters, controlling the block sparsity.
This formula composes the encoder of our proposed deep neural network.

It is worth mentioning that setting $\{\bv_i\}_{i=1}^N$ as learnable parameters is crucial because 
in previous \nrsfm{} algorithms --  low-rank~\cite{dai2014simple},
union-of-subspaces~\cite{zhu2014complex}, or block-sparsity~\cite{kong2016prior} priors -- the weight 
associated with shape regularization (\eg low-rank or sparsity) is determined through a cumbersome and 
slow grid-search process. In our approach, this weighting is learned simultaneously with all other
parameters, removing the need for irksome cross-validation.

\subsection{Code and Camera Recovery}
\label{sec:camera}
Recall that in Section~\ref{sec:hbsc}, we define $\Psiv = \psiv\otimes\Mv$. By denoting the $k$-th 
block in $\Psiv_N$ as $\Psiv_N^k$ and the $k$-th element in $\psiv_N$ as $\psiv_N^k$. we have
\begin{equation}
    \Psiv_N^k = \psiv_N^k \Mv.
\end{equation}
Now, we want to estimate the regular sparse hidden code~$\psiv_N$ and camera~$\Mv$ given $\Psiv_N$. 
It is obvious that if one of them is known beforehand, then the other one can be solved easily. For 
example, if $\Mv$ is known, then $\psiv_N^k$ can be estimated by
\begin{equation}
    \psiv_N^k = \frac{1}{6}\sum_{i=1}^3\sum_{j=1}^2\frac{[\Psiv_N^k]_{ij}}{[\Mv]_{ij}} 
              = \sum_{i=1}^3\sum_{j=1}^2 \frac{1}{6[\Mv]_{ij}}[\Psiv_N^k]_{ij},
    \label{eq:old_psiv}
\end{equation}
where $[\cdot]_{ij}$ denotes the $ij$-th element in the argument matrix.
Note that actually a single element in camera M and its correspondence in $\Psiv_N$ are sufficient 
to estimate the scaler $\psiv_N^k$, but, for robust estimation, an average over all elements 
($3\times2$ block results in totally 6 elements) is utilized here. 
Further, if $\psiv_N$ is known, then $\Mv$ can be estimated by
\begin{equation}
    \Mv = \frac{1}{K_N}\sum_{k=1}^{K_N} \frac{\Psiv_N^k}{\psiv_N^k} 
        = \sum_{k=1}^{K_N} \frac{1}{K_N\psiv_N^k} \Psiv_N^k.
    \label{eq:old_M}
\end{equation}
Note that a single element in $\psiv_N$ and a corresponding block in $\Psiv_N$ is again sufficient to
estimate $\Mv$ but, for robustness, we utilize an average across all blocks.

In the literature of the field~\cite{dai2014simple, akhter2011trajectory, kong2016prior}, these two 
coupled variables are mainly solved by a carefully designed algorithm that utilizes the orthonormal
constraint to solve the camera first and then the sparse hidden code.  However, this heuristic is 
quite fragile and it is even worse when the estimation of $\Psiv_N$ is bothered by noise. Further, 
it has difficulty deciding the sign ambiguity of each sparse code.
In this paper, we propose a novel relaxation, decoupling equations~(\ref{eq:old_psiv}) 
and~(\ref{eq:old_M}) by introducing two learnable parameters $\beta$ and $\gamma$, 
specifically, 
\begin{equation}
    \psiv_N^k = \sum_{i=1}^3 \sum_{j=1}^2 \beta_{ij}[\Psiv_N^{k}]_{ij},
\end{equation}
\begin{equation}
    \Mv = \sum_{k=1}^{K_N} \gamma_k \Psiv_N^k.
    \label{eq:estimate_camera}
\end{equation}
It is clear that $\psiv$ and $\Mv$ are intrinsically linked -- but our proposed approach seems to ignore this dependency. We resolve this inconsistency, however, by enforcing an orthonormal constraint for the camera in our loss function shown in Section~\ref{sec:loss}.
This relaxation has the further advantage of eliminating fragile heuristics and giving substantial computational savings. \figurename~\ref{fig:architecture} represents this process via convolutions for conciseness and descent 
visualization. 

\subsection{Decoder}
Given the sparse hidden code $\psiv_N$ and hierarchical dictionaries $\{\Dv_i\}_{i=1}^N$, the 3D shape 
vector $\sv$ could be recovered via (\ref{eq:mlsc}). In practice, instead of forming a purely linear 
decoder, we preserve soft-thresholding in each layer. This non-linear decoder is expected to further 
enforce sparsity and improve robustness.  Formally,
\begin{equation}
 \begin{aligned}
  \psiv_{N-1} & = \eta(\Dv_N \psiv_N; \bv_N'), \\
              & \quad \vdots                     \\
  \psiv_1     & = \eta(\Dv_2 \psiv_2; \bv_2'), \\
  \sv         & = \Dv^\sharp_1\psiv_1.
 \end{aligned}
\end{equation}
This portion forms the decoder of our deep neural network.

\subsection{Loss Function}
\label{sec:loss}
Until now, the 3D shape $\Sv$ is estimated via the proposed encoder and decoder architecture given the
hierarchical dictionaries, which is denoted as $\Sc\big(\Wv \vert \{\Dv_i\}_{i=1}^N\big)$ for simplicity.
Further, the camera $\Mv$ is also estimated via the encoder and a linear combination given the dictionaries, 
which is denoted as $\Mc\big( \Wv \vert \{\Dv_i\}_{i=1}^N\big)$. Our loss function is thus defined as
\begin{equation}
    \begin{aligned}
        \min_{\{\Dv\}_{i=1}^N} &\sum_{\Wv \in \mathbb{W}} \left\Vert \Wv - \Sc\big(\Wv \vert \{\Dv_i\}_{i=1}^N\big)\Uv \Vv^{T} \right\Vert_F\\
        \st &\Uv\Sigma\Vv^T = \Mc\big(\Wv \vert \{\Dv_i\}_{i=1}^N\big),
    \end{aligned}
\end{equation}
which is the summation of reprojection error. To ensure the success of the orthonormal constraint on
the camera, we introduce the Singular Value Decomposition~(SVD) to hard code the singular value of $\Mv$
to be exact ones. As mentioned in Section~\ref{sec:camera}, reprojecting the estimated 3D shape via the 
estimated camera (\ie{} left multiplying $\Mv$ to $\Sv$) implicitly re-build the bonds between the camera 
$\Mv$ and the sparse hidden code $\psiv_N$ (in the form of 3D shape $\Sv$).

\subsection{Implementation Issues}
The Kronecker product of identity matrix $\Iv_3$ dramatically increases the
time and space complexity of our approach. To eliminate it and make parameter
sharing easier in modern deep-learning environments~(\eg~TensorFlow, PyTorch),
we reshape the filters and features so that the matrix multiplication in each step can be equivalently computed via multi-channel convolution~($*$) and transposed convolution~($*^T$).
We first reshape the 2D input correspondences $\Wv$ into a three-dimensional tensor 
$\wsf \in \RR^{1\times 2\times P}$, which can be considered in the deep-learning community as a 
$1\times 2$ image with $P$ channels. Then, we reshape the first dictionary $\Dv_1^\sharp$ into a
four-dimensional tensor $\dsf_1^\sharp \in \RR^{3\times1\times K_1 \times P}$, which can 
be interpreted as a convolutional kernel in size $3\times 1$ with $K_1$ input channels and $P$ output 
channels. Therefore, we have
\begin{equation}
 (\Dv_1^\sharp)^T\Wv = \dsf_1^\sharp *^T \wsf,
\end{equation}
which helps us to maintain a uniform dictionary shape and is consequently easier to share parameters.
We then reshape each dictionary $\Dv_{i}$ other than the first one into a four-dimensional 
tensor $\dsf_{i} \in \RR^{1\times 1\times K_{i} \times K_{i-1}}$ and the hidden 
block sparse code $\Psiv_{i}$ into a three-dimensional tensor $\Psi_i\in\RR^{3\times 2\times K_i}$.
Therefore, we have
\begin{equation}
 (\Dv_{i+1}\otimes\Iv_3)^T\Psiv_i = \dsf_{i+1} *^T \Psi_{i},
\end{equation}
which helps us to eliminate the Kronecker product.
Finally, based on the above reshape, the dictionary-code multiplication is simplified as
\begin{equation}
 \Dv_i\psiv_i = \dsf_{i} * \psi_{i}.
\end{equation}

As for the architecture design, we only control three hyper parameters: 1) the number of dictionaries
$N$, 2) the number of atoms in the first dictionary $K_1$, and 3) the number of atoms in the last 
dictionary $K_N$. We then linearly sample $K_2, \dots, K_{N-1}$ between $K_1$ and $K_N$. As for training,
we implement our neural network via TensorFlow and train it using an Adam optimizer with a learning rate 
exponentially decayed from 0.001.

\subsection{Replacing Soft-thresholding via ReLU}
Recall in Section~\ref{sec: scdnn}, Papyan~\etal{} replaced the soft-thresholding operator $\eta$ by ReLU as a result of the non-negativity constraint. Actually, it can easily be demonstrated that a linear 
(block) sparse model can always be transferred equivalently to a model only using non-negative (block)
sparse code \ie
\begin{equation}
\label{eq:non-negativity}
\Wv = \Dv\Psiv = \begin{bmatrix}\Dv & -\Dv\end{bmatrix} \begin{bmatrix}\Psiv^{+} \\ -\Psiv^{-}\end{bmatrix}.
\end{equation}
where $\Psiv^{+}$ and $\Psiv^{-}$ are positive and negative parts of $\Psiv$ respectively and
$\Psiv^{+} + \Psiv^{-} = \Psiv$. The concatenation of $\Psiv^{+}$ and $-\Psiv^{-}$ is still block sparse
and now becomes non-negative.
From this observation, we introduce the non-negativity constraints without the loss of generality and 
relax the dictionaries so that they are not bothered by mirrored structures.  Interestingly, our 
proposed method on estimating cameras in (\ref{eq:estimate_camera}) is compatible with the change, \ie
\begin{equation}
    \Mv = \sum_{k=1}^{K_N} \gamma_k\Psiv_N^k = \sum_{k=1}^{K_N} \gamma_k (\Psiv_N^k)^+ \sum_{k=1}^{K_N} -\gamma_k (-\Psiv_N^k)^-.
\end{equation}
All of these enable us to utilize ReLU to replace the soft-thresholding.  ReLU is good because it is 
closer to deep learning packages while soft-thresholding is more compact in size of parameters.  
An experiment comparing between soft-thresholding and ReLU is in the Appendix. It is demonstrated
that no discernible difference in the accuracy of reconstructions is observed. 
Therefore, we decide to use ReLU for the remaining sections and experiments, making our 
approach closer to leading techniques in deep learning and more accessible and approachable to the 
public.  

\section{Occlusion and Weak Perspective}
\label{sec:weak-perspective}

\subsection{Occlusion}
It is commonly observed in real images that a certain portion of key points are occluded by
other objects or the object itself. For example, we typically see two wheels of a sedan instead of four.
An often-used strategy is to recover the missing entries in $\Wv$ by matrix completion before feeding
it into the proposed pipeline. A commonly used shape prior for matrix completion is low-rank, even
for some union-of-subspaces algorithms~\cite{agudo2018image}. This is problematic.

In this paper, we derive a solution from the ISTA to handle missing entries, which turns out as a 
quite simple but well-functioning operation. We observe that missing entries break the first layer 
of encoder but once $\Psiv_1$ is estimated, all other layers can execute with no trouble. 
Based on this observation, we first introduce a diagonal matrix $\Omegav\in\RR^{P\times P}$, whose 
element on the main diagonal is zero if the corresponding point in $\Wv$ is missing; otherwise, one 
and all other elements except diagonal are zeros. 
With the help of the mask $\Omegav$, the objective function w.r.t the first layer is
\begin{equation}
    \min_{\Psiv_1} \Vert \Omegav(\Wv - \Dv_1^\sharp\Psiv_1 )\Vert_F^2 \quad \text{s.t.}~\Vert \Psiv_1\Vert_0^{(3\times 2)} < \lambda_1.
\end{equation}
Following the same derivation in Section~\ref{sec:architecture}, a masked ISTA is to iteratively
execute the following two steps:
\begin{gather}
 \Vv = \Psiv_1^{[i]} - \alpha(\Dv_1^\sharp)^T\Omegav^T\Omegav(\Dv_1^\sharp\Psiv_1^{[i]} - \Wv), \\
 \Psiv_1^{[i+1]} = \argmin_{\Uv} \frac{1}{2}\Vert \Uv - \Vv \Vert^2_F + \tau\Vert\Uv\Vert_{F1}^{(3\times2)},
\end{gather}
By (\ref{eq:block-relu}), it is implied that the single-iteration block ISTA with mask is
\begin{equation}
    \Psiv_1 = \eta\big((\Dv_1^\sharp)^T\Omegav\Wv - \bv\otimes\Iv_{3\times 2}\big).
\end{equation}
This is equivalently to set missing entries to zero and then feed into the proposed deep neural network.

\subsection{Scale and Translation}
The main difference between weak perspective and orthogonal projection is additional scale
and translation besides rotation. Due to the ambiguity between camera scale and 3D shape size, 
we do not solve the camera scale explicitly, but consider the scale to be one and reconstruct a scaled
3D shape. To alleviate the effect of the scale on optimization, we normalize the 2D correspondences into
a unit bounding box before feeding into the proposed neural network.

Translation is not a problem and can even be eliminated when all points are visible. This is because one can
always remove the camera translation by shifting the center of 2D correspondences to the image origin.
However, this is not true when some correspondences are missing. Formally, $i\in \Omega$ denotes that the $i$-th point is visible and $(u_i, v_i)$ is the image coordinate of the $i$-th point. Shifting the center of 
all points (where missing entries are set to zero) to the origin remains a translation residual
\begin{equation}
\frac{1}{n}\sum_{i} \begin{bmatrix} u_i \\ v_i\end{bmatrix} - \frac{1}{n}\sum_{i\in\Omega}\begin{bmatrix} u_i \\ v_i\end{bmatrix} = \frac{1}{n}\sum_{i\not\in\Omega}\begin{bmatrix} u_i \\ v_i\end{bmatrix}.
\end{equation}
When key points distribute closely in a cluster and a small portion of them are missing, the residual 
translation could be treated as some sort of noise perturbation and consequently need no further operation.
Otherwise, we need to solve the translation explicitly.

Consider the camera projection with translation $\tv$, \ie
\begin{equation}
 \Wv = \begin{bmatrix} u_1 & v_1 \\ u_2 & v_2 \\ \vdots & \vdots \\ u_P & v_P \end{bmatrix} = 
 \begin{bmatrix} x_1 & y_1 & z_1 & 1\\ x_2 & y_2 & z_2 & 1\\ \vdots & \vdots & \vdots & \vdots \\ x_P & y_P & z_P  & 1\end{bmatrix} \begin{bmatrix} \Mv \\ \tv^T \end{bmatrix}.
\end{equation}
We could introduce an auxiliary variable $\epsilon$ 
\begin{equation}
 \Wv =  
 \begin{bmatrix} x_1 & y_1 & z_1 & \epsilon\\ x_2 & y_2 & z_2 & \epsilon\\ \vdots & \vdots & \vdots & \vdots\\ x_P & y_P & z_P  & \epsilon\end{bmatrix} \begin{bmatrix} \Mv \\ \tv^T/\epsilon \end{bmatrix}=\tilde{\Sv} \tilde{\Mv}
\end{equation}
such that $\tilde{\Sv}$ satisfies the proposed hierarchical sparse model in (\ref{eq:mlbsc_s}) after 
appending ones to each dictionary. Therefore, a similar neural network could be derived from a 4-by-2
block sparse ISTA as $\tilde{\Mv}\in\RR^{4\times2}$.

\begin{figure*}[!t]
    \centering
    \includegraphics[width=0.12\linewidth]{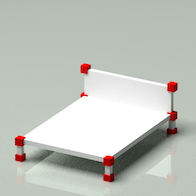}
    \includegraphics[width=0.12\linewidth]{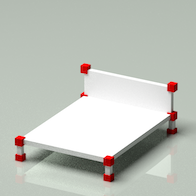}
    \includegraphics[width=0.12\linewidth]{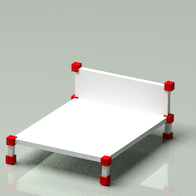}
    \includegraphics[width=0.12\linewidth]{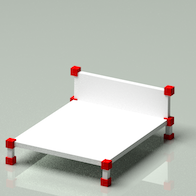}
    \includegraphics[width=0.12\linewidth]{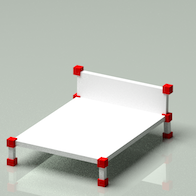}
    \includegraphics[width=0.12\linewidth]{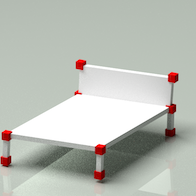}
    \includegraphics[width=0.12\linewidth]{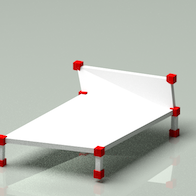}
    \includegraphics[width=0.12\linewidth]{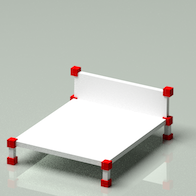}
    
    \vspace{1mm}
    
    \includegraphics[width=0.12\linewidth]{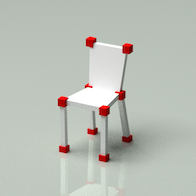}
    \includegraphics[width=0.12\linewidth]{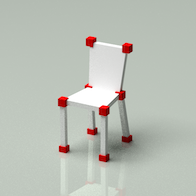}
    \includegraphics[width=0.12\linewidth]{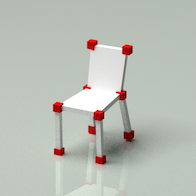}
    \includegraphics[width=0.12\linewidth]{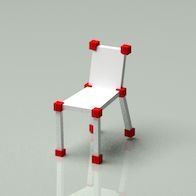}
    \includegraphics[width=0.12\linewidth]{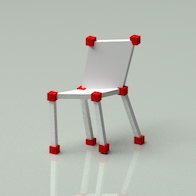}
    \includegraphics[width=0.12\linewidth]{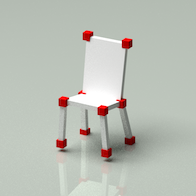}
    \includegraphics[width=0.12\linewidth]{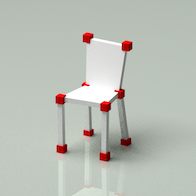}
    \includegraphics[width=0.12\linewidth]{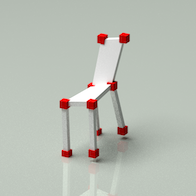}
    
    \vspace{1mm}
    
    \includegraphics[width=0.12\linewidth]{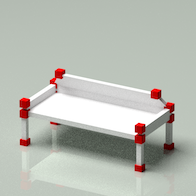}
    \includegraphics[width=0.12\linewidth]{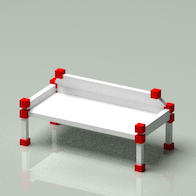}
    \includegraphics[width=0.12\linewidth]{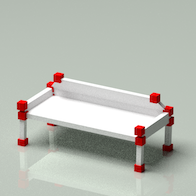}
    \includegraphics[width=0.12\linewidth]{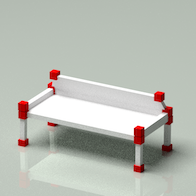}
    \includegraphics[width=0.12\linewidth]{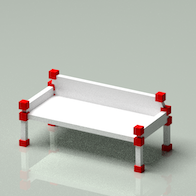}
    \includegraphics[width=0.12\linewidth]{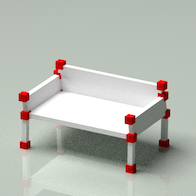}
    \includegraphics[width=0.12\linewidth]{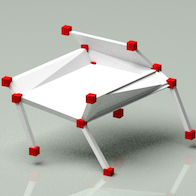}
    \includegraphics[width=0.12\linewidth]{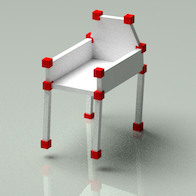}
    
    \vspace{1mm}
    
    \includegraphics[width=0.12\linewidth]{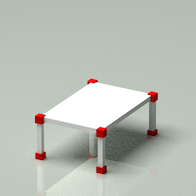}
    \includegraphics[width=0.12\linewidth]{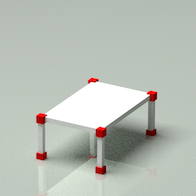}
    \includegraphics[width=0.12\linewidth]{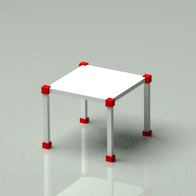}
    \includegraphics[width=0.12\linewidth]{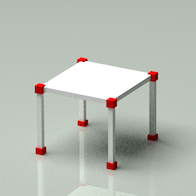}
    \includegraphics[width=0.12\linewidth]{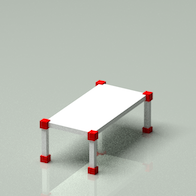}
    \includegraphics[width=0.12\linewidth]{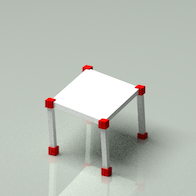}
    \includegraphics[width=0.12\linewidth]{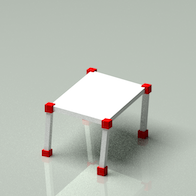}
    \includegraphics[width=0.12\linewidth]{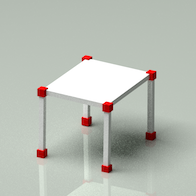}
    \caption{Qualitative evaluation on IKEA dataset. From top to bottom are tables, chairs, sofas and tables.
    From left to right are ground-truth and respectively reconstructions by ours, 
    RIKS~\cite{hamsici2012learning}, KSTA~\cite{gotardo2011kernel}, NLO~\cite{del2007non},
    SFC~\cite{kong2016sfc}, CNS~\cite{lee2016consensus}, BMM~\cite{dai2014simple}. In each rendering, 
    red cubes are reconstructed points but the planes and bars are manually added for descent visualization.}
    \label{fig:ikea}
\end{figure*}

\section{Experiments}
We conduct extensive experiments to evaluate the performance of our deep
solution to solving \nrsfm{} and \sfc{} problems. For quantitative evaluation,
we follow the metric normalized mean 3D error reported
in~\cite{akhter2009nonrigid, dai2014simple, gotardo2011kernel, agudo2018image}.
Our implementation, processed data, and pre-trained models are publicly accessible for future
comparison\footnote{https://github.com/kongchen1992/deep-nrsfm}.

\subsection{IKEA Furniture}
We first apply our method to a furniture dataset, IKEA dataset~\cite{lpt2013ikea,
 wu2016single}. The IKEA dataset contains four object categories: bed, chair,
sofa, and table. For each object category, we project the 3D ground-truth by the orthogonal
cameras annotated from real images. Since fully annotated images are limited, we thereby augment 
them with 2,000 projections under randomly generated orthogonal cameras. 
The errors are evaluated only on frames using cameras from real images. Numbers are summarized into
Table~\ref{tab:sfc}. One can observe that our method outperforms baselines
in the order of magnitude, clearly showing the superiority of our model.
For qualitative evaluation, we randomly select a frame from each object category
and show these frames in \figurename~\ref{fig:ikea} against ground-truth and baselines. As shown, 
our reconstructed landmarks effectively depict the 3D geometry of objects
and our method is able to cover subtle geometric details.

\begin{table}[!h]
\renewcommand{\arraystretch}{1.3}
\caption{Quantitative Comparison against State-Of-The-Art Algorithms using IKEA Dataset in Normalized 3D Error.}
\label{tab:sfc}
\centering
\begin{tabular}{r|cccccc}
\hline\hline
Furnitures & Bed & Chair & Sofa & Table & Average        & Relative      \\\hline
KSTA~\cite{gotardo2011kernel}       & 0.069                       & 0.158                         & 0.066                        & 0.217                         & 0.128          & 12.19         \\
BMM~\cite{dai2014simple}        & 0.059                       & 0.330                         & 0.245                        & 0.211                         & 0.211          & 20.12         \\
CNR~\cite{lee2016consensus}        & 0.227                       & 0.163                         & 0.835                        & 0.186                         & 0.352          & 33.55         \\
NLO~\cite{del2007non}        & 0.245                       & 0.339                         & 0.158                        & 0.275                         & 0.243          & 23.18         \\
RIKS~\cite{hamsici2012learning}       & 0.202                       & 0.135                         & 0.048                        & 0.218                         & 0.117          & 11.13         \\
SPS~\cite{kong2016prior}        & 0.971                       & 0.946                         & 0.955                        & 0.280                         & 0.788          & 74.96         \\
SFC~\cite{kong2016sfc}        & 0.247                       & 0.195                         & 0.233                        & 0.193                         & 0.217          & 20.67         \\\hline
OURS       & \textbf{0.004}              & \textbf{0.019}                & \textbf{0.005}               & \textbf{0.012}                & \textbf{0.010} & \textbf{1.00} \\\hline\hline
\end{tabular}
\end{table}

\subsection{PASCAL3D+ Dataset}
We then apply our method to the PASCAL3D+ dataset~\cite{xiang2014beyond}, which contains
twelve object categories. Following the experiment setting reported in~\cite{agudo2018image},
we also utilize eight categories: aeroplane, bicycle, bus, car, chair, dining table,
motorbike and sofa.  To explore the performance in various situations, we design experiments with
respect to
\begin{itemize}
    \item Orthogonal or weak perspective projection?
    \item Complete or missing measurement?
    \item Clean data or Gaussian noise perturbed?
\end{itemize}
Totally, there are eight configurations. Specifically, for projection setting, we randomly generate rotation 
matrices for orthogonal projection while additionally utilizing random scale and random translation for weak 
perspective projection.  For missing data, we randomly sample approximately $10\%$ of
points missing for each category. For noise, we corrupt 2D correspondences with a zero mean 
Gaussian perturbation, following the same noise ratio in~\cite{agudo2018image}.
For the translation residual, we simply treat it as noise and handle it with a 3-by-2 block sparse model. 
In Table~\ref{tab:pascal}, we report the normalized mean 3D error of our proposed method and
state-of-the-arts: KSTA~\cite{gotardo2011kernel}, RIKS~\cite{hamsici2012learning}, 
CNS~\cite{lee2016consensus}, NLO~\cite{del2007non}, SFC~\cite{kong2016sfc}, SPS~\cite{kong2016prior},
and BMM~\cite{dai2014simple}. For readers' interest, one can compare our numbers against the Table 2
in~\cite{agudo2018image} for more baselines. 

\newcommand{\PreserveBackslash}[1]{\let\temp=\\#1\let\\=\temp}
\newcolumntype{C}{>{\PreserveBackslash\centering}p{6.3mm}}
\begin{table*}[!t]
\renewcommand{\arraystretch}{1.3}
\caption{Quantitative Comparison against State-Of-The-Art Algorithms using PASCAL3D Dataset. In each configuration, numbers from top to bottom are for category aeroplane, bicycle, bus, car, chair, diningtable
motorbike and sofa.}
\label{tab:pascal}
\centering
\begin{tabular}{|l|l|CCCCCCCC|CCCCCCCC|}
\hline\hline
\multicolumn{2}{|c|}{\multirow{2}{*}{}}     & \multicolumn{8}{c|}{No Noise Perturbation}                     & \multicolumn{8}{c|}{Added Noise Perturbation}                  \\
\cline{3-18}
\multicolumn{2}{|c|}{}                      & OURS  & KSTA  & RIKS  & CNS   & NLO   & SFC   & SPS   & BMM   & OURS  & KSTA  & RIKS  & CNS   & NLO   & SFC   & SPS   & BMM   \\
\hline
\parbox[t]{2mm}{\multirow{16}{*}{\rotatebox[origin=c]{90}{Orthogonal Projection}}} & \parbox[t]{2mm}{\multirow{8}{*}{\rotatebox[origin=c]{90}{Complete Measurement}}}   & \textbf{0.013} & 0.161 & 0.562 & 0.636 & 0.175 & 0.499 & 0.902 & 1.030 & \textbf{0.026} & 0.175 & 0.583 & 0.626 & 0.167 & 0.518 & 0.761 & 1.177 \\
& & \textbf{0.003} & 0.249 & 0.826 & 0.732 & 0.285 & 0.370 & 0.959 & 1.247 & \textbf{0.009} & 0.253 & 0.779 & 0.715 & 0.916 & 0.367 & 1.065 & 1.424 \\
& & \textbf{0.004} & 0.201 & 0.578 & 0.443 & 0.262 & 0.255 & 0.902 & 0.728 & \textbf{0.012} & 0.196 & 0.450 & 0.442 & 0.320 & 0.253 & 1.096 & 0.754 \\
& & \textbf{0.003} & 0.124 & 0.497 & 0.497 & 0.135 & 0.284 & 0.955 & 1.006 & \textbf{0.012} & 0.162 & 0.557 & 0.496 & 0.192 & 0.285 & 0.879 & 0.915 \\
& & \textbf{0.009} & 0.191 & 0.748 & 0.540 & 0.145 & 0.223 & 1.018 & 1.381 & \textbf{0.028} & 0.190 & 0.668 & 0.554 & 0.107 & 0.224 & 0.927 & 1.251 \\
& & \textbf{0.030} & 0.244 & 0.778 & 0.549 & 0.234 & 0.220 & 0.707 & 1.351 & \textbf{0.040} & 0.238 & 0.721 & 0.521 & 0.450 & 0.219 & 0.968 & 1.420 \\
& & \textbf{0.001} & 0.254 & 0.703 & 0.647 & 0.320 & 0.356 & 1.090 & 1.033 & \textbf{0.004} & 0.251 & 0.722 & 0.629 & 0.168 & 0.366 & 0.938 & 1.029 \\
& & \textbf{0.007} & 0.401 & 0.798 & 0.623 & 0.055 & 0.302 & 0.779 & 1.017 & \textbf{0.020} & 0.333 & 0.725 & 0.627 & 0.064 & 0.297 & 1.041 & 1.315 \\
\cline{2-18}
&\parbox[t]{0.5mm}{\multirow{8}{*}{\rotatebox[origin=c]{90}{Missing Measurement}}} & \textbf{0.033} & 0.533 & 0.515 & 0.693 & 0.348 & 0.496 & 1.076 & 1.154 & \textbf{0.065} & 0.434 & 0.514 & 0.707 & 0.382 & 0.493 & 0.815 & 1.199 \\
& & \textbf{0.021} & 0.584 & 0.540 & 0.854 & 0.106 & 0.376 & 1.112 & 1.372 & \textbf{0.028} & 0.566 & 0.560 & 0.835 & 0.459 & 0.372 & 1.201 & 1.286 \\
& & \textbf{0.018} & 0.357 & 0.316 & 0.517 & 0.317 & 0.254 & 1.273 & 0.728 & \textbf{0.057} & 0.364 & 0.323 & 0.526 & 0.079 & 0.245 & 0.791 & 0.743 \\
& & \textbf{0.010} & 0.400 & 0.334 & 0.598 & 0.089 & 0.286 & 0.918 & 1.014 & \textbf{0.023} & 0.391 & 0.299 & 0.587 & 0.111 & 0.285 & 1.077 & 1.244 \\
& & \textbf{0.024} & 0.599 & 0.581 & 0.601 & 0.102 & 0.228 & 1.184 & 1.242 & \textbf{0.066} & 0.571 & 0.479 & 0.593 & 0.103 & 0.229 & 1.153 & 1.274 \\
& & \textbf{0.040} & 0.554 & 0.473 & 0.602 & 0.171 & 0.224 & 1.264 & 1.414 & \textbf{0.050} & 0.494 & 0.408 & 0.587 & 0.177 & 0.228 & 1.019 & 1.098 \\
& & \textbf{0.009} & 0.539 & 0.501 & 0.729 & 0.177 & 0.366 & 0.892 & 1.117 & \textbf{0.032} & 0.523 & 0.528 & 0.730 & 0.154 & 0.363 & 1.100 & 1.157 \\
& & \textbf{0.015} & 0.573 & 0.567 & 0.728 & 0.911 & 0.301 & 1.214 & 1.171 & \textbf{0.039} & 0.576 & 0.590 & 0.727 & 0.080 & 0.307 & 1.252 & 1.017\\
\hline
\parbox[t]{2mm}{\multirow{16}{*}{\rotatebox[origin=c]{90}{Weak Perspective Projection}}} & \parbox[t]{2mm}{\multirow{8}{*}{\rotatebox[origin=c]{90}{Complete Measurement}}}   & \textbf{0.034} & 0.402 & 0.460 & 0.667 & 0.192 & 0.500 & 1.123 & 1.055 & \textbf{0.046} & 0.525 & 0.489 & 0.644 & 0.206 & 0.527 & 0.961 & 1.203 \\
&& \textbf{0.008} & 0.576 & 0.817 & 0.707 & 0.595 & 0.373 & 1.172 & 1.301 & \textbf{0.029} & 0.618 & 0.729 & 0.760 & 0.930 & 0.368 & 1.202 & 1.331 \\
&& \textbf{0.017} & 0.480 & 0.582 & 0.458 & 0.205 & 0.251 & 1.380 & 0.743 & \textbf{0.044} & 0.384 & 0.443 & 0.443 & 0.666 & 0.248 & 0.820 & 0.739 \\
&& \textbf{0.015} & 0.369 & 0.573 & 0.504 & 0.175 & 0.284 & 1.090 & 1.051 & \textbf{0.022} & 0.409 & 0.475 & 0.524 & 0.178 & 0.285 & 0.836 & 1.342 \\
&& \textbf{0.013} & 0.621 & 0.832 & 0.540 & 0.197 & 0.224 & 0.970 & 1.220 & \textbf{0.026} & 0.497 & 0.622 & 0.543 & 0.122 & 0.226 & 1.283 & 1.284 \\
&& \textbf{0.025} & 0.647 & 0.829 & 0.533 & 0.428 & 0.220 & 0.927 & 1.447 & \textbf{0.068} & 0.585 & 0.629 & 0.506 & 0.303 & 0.220 & 0.993 & 1.123 \\
&& \textbf{0.003} & 0.614 & 0.739 & 0.662 & 0.180 & 0.359 & 1.406 & 1.069 & \textbf{0.018} & 0.607 & 0.789 & 0.671 & 0.159 & 0.362 & 1.101 & 1.019 \\
&& \textbf{0.022} & 0.609 & 0.792 & 0.632 & 0.070 & 0.295 & 0.976 & 0.980 & \textbf{0.041} & 0.606 & 0.684 & 0.644 & 0.062 & 0.301 & 1.603 & 1.165 \\
\cline{2-18}
&\parbox[t]{2mm}{\multirow{8}{*}{\rotatebox[origin=c]{90}{Missing Measurement}}}& \textbf{0.102} & 0.461 & 0.531 & 0.727 & 0.670 & 0.502 & 1.162 & 1.150 & \textbf{0.157} & 0.449 & 0.571 & 0.737 & 0.742 & 0.493 & 0.984 & 1.220 \\
&&\textbf{0.048} & 0.499 & 0.572 & 0.875 & 0.115 & 0.372 & 1.312 & 1.279 & \textbf{0.084} & 0.668 & 0.708 & 0.895 & 0.141 & 0.375 & 1.003 & 1.405 \\
&&\textbf{0.066} & 0.356 & 0.341 & 0.553 & 0.091 & 0.250 & 0.912 & 0.752 & \textbf{0.091} & 0.383 & 0.365 & 0.557 & 0.139 & 0.253 & 0.985 & 0.752 \\
&&\textbf{0.027} & 0.402 & 0.403 & 0.637 & 0.093 & 0.280 & 0.949 & 0.954 & \textbf{0.081} & 0.355 & 0.358 & 0.619 & 0.109 & 0.293 & 1.023 & 1.063 \\
&&\textbf{0.077} & 0.484 & 0.485 & 0.607 & 0.118 & 0.227 & 1.107 & 1.263 & \textbf{0.122} & 0.522 & 0.434 & 0.601 & 0.123 & 0.224 & 1.037 & 1.263 \\
&&\textbf{0.091} & 0.463 & 0.465 & 0.594 & 0.174 & 0.232 & 1.210 & 1.229 & \textbf{0.136} & 0.558 & 0.528 & 0.612 & 0.173 & 0.225 & 1.151 & 1.510 \\
&&\textbf{0.056} & 0.561 & 0.656 & 0.779 & 0.201 & 0.367 & 1.119 & 1.125 & \textbf{0.051} & 0.544 & 0.585 & 0.763 & 0.191 & 0.369 & 1.039 & 1.017 \\
&&\textbf{0.066} & 0.529 & 0.615 & 0.728 & 0.081 & 0.311 & 1.730 & 1.150 & \textbf{0.082} & 0.543 & 0.548 & 0.730 & 0.156 & 0.299 & 0.890 & 1.146 \\
\hline\hline
\end{tabular}%
\end{table*}

\begin{figure*}[!t]
    \centering
    \includegraphics[width=0.12\linewidth]{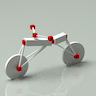}
    \includegraphics[width=0.12\linewidth]{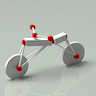}
    \includegraphics[width=0.12\linewidth]{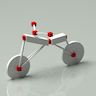}
    \includegraphics[width=0.12\linewidth]{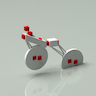}
    \includegraphics[width=0.12\linewidth]{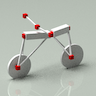}
    \includegraphics[width=0.12\linewidth]{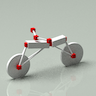}
    \includegraphics[width=0.12\linewidth]{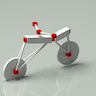}
    \includegraphics[width=0.12\linewidth]{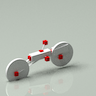}
    
    \vspace{1mm}
    
    \includegraphics[width=0.12\linewidth]{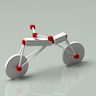}
    \includegraphics[width=0.12\linewidth]{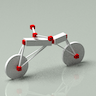}
    \includegraphics[width=0.12\linewidth]{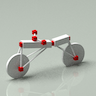}
    \includegraphics[width=0.12\linewidth]{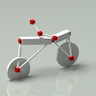}
    \includegraphics[width=0.12\linewidth]{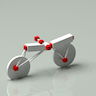}
    \includegraphics[width=0.12\linewidth]{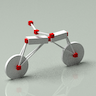}
    \includegraphics[width=0.12\linewidth]{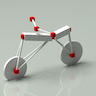}
    \includegraphics[width=0.12\linewidth]{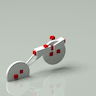}
    
    \vspace{1mm}
    
    \includegraphics[width=0.12\linewidth]{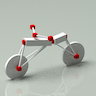}
    \includegraphics[width=0.12\linewidth]{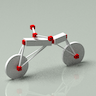}
    \includegraphics[width=0.12\linewidth]{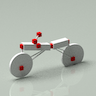}
    \includegraphics[width=0.12\linewidth]{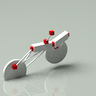}
    \includegraphics[width=0.12\linewidth]{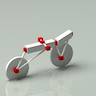}
    \includegraphics[width=0.12\linewidth]{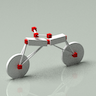}
    \includegraphics[width=0.12\linewidth]{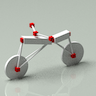}
    \includegraphics[width=0.12\linewidth]{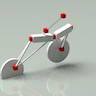}
    
    \vspace{1mm}
    
    \includegraphics[width=0.12\linewidth]{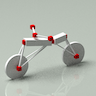}
    \includegraphics[width=0.12\linewidth]{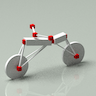}
    \includegraphics[width=0.12\linewidth]{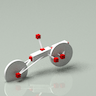}
    \includegraphics[width=0.12\linewidth]{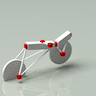}
    \includegraphics[width=0.12\linewidth]{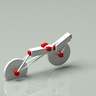}
    \includegraphics[width=0.12\linewidth]{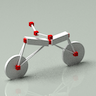}
    \includegraphics[width=0.12\linewidth]{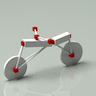}
    \includegraphics[width=0.12\linewidth]{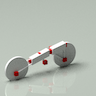}
    
    \caption{Qualitative evaluation on PASCAL3D dataset. From top to bottom are configurations
    1) orthogonal projection with no missing points, 2) orthogonal projection with missing points,
    3) weak perspective projection with no missing points, 4) weak perspective projection with missing points.
    All these four configurations are perturbed by Gaussian noise. From left to right are
    ground-truth, ours, KSTA~\cite{gotardo2011kernel}, RIKS~\cite{hamsici2012learning}, 
    CNS~\cite{lee2016consensus}, NLO~\cite{del2007non}, SFC~\cite{kong2016sfc}, SPS~\cite{kong2016prior}. 
    In each rendering of reconstruction, red cubes are reconstructed points but the planes and bars are
    manually added for visualization. }
    \label{fig:pascal}
\end{figure*}

From Table~\ref{tab:pascal}, one can observe that our proposed method achieves considerably more accurate
reconstructions for all cases, and for some cases, more than ten times the amount of smaller 3D errors than
state-of-the-arts. It clearly demonstrates the high precision of our proposed deep neural network. 
By comparing between clean and noisy configurations, it is shown that our proposed method has high 
robustness, where our method applied on noisy data even outperforms state-of-the-arts on clean data. 
By comparing between orthogonal and weak perspective projections, it is demonstrated that our proposed 
3-by-2 block sparse model can handle scale and translation properly, even with missing data. In the 
configuration with missing measurement, KSTA, RIKS, BMM, CNS, and SPS  use the matrix completion 
algorithm proposed by~\cite{gotardo2011computing} to recover missing entries first, but our proposed 
method, SFC, and NLO can directly optimize over partially-visible 2D measurements, which are more 
capable at handling missing data. This is verified 
by Table~\ref{tab:pascal}, where OURS, SFC, and NLO sacrifice less performance than others when 
handling missing data. For qualitative evaluation, we use ``motorbike'' as an exemplar category and 
randomly select a frame from four configurations:
1) orthogonal+complete+noise, 
2) orthogonal+missing+noise, 
3) weak perspective+complete+noise, and
4) weak perspective+missing+noise, showing in \figurename~\ref{fig:pascal}. One can observe that our
proposed method outperforms KSTA, RIKS, CNS and SPS obviously and beat NLO and SFC in reconstruction
details, \eg~handlebar. The figure also verifies that KSTA, RIKS, CNS, and SPS break easily with missing
points while ours, SFC, and NLO maintain a nice stability against missing entries.

\begin{table}[!t]
\renewcommand{\arraystretch}{1.3}
\caption{Quantitative Comparison aginst State-Of-The-Arts using CMU Motion Capture Dataset in Normalized 3D Error}
\label{tab:mocap}
\centering
\begin{tabular}{r|cccc|c}
\hline\hline
SUBJECT & OURS  & CNS   & NLO   & SPS   & UNSEEN \\
\hline
1   & \textbf{0.176} & 0.613 & 1.218 & 1.282 & 0.362  \\
5   & \textbf{0.221} & 0.657 & 1.160 & 1.122 & 0.331  \\
18  & \textbf{0.082} & 0.542 & 0.917 & 0.954 & 0.438  \\
23  & \textbf{0.054} & 0.604 & 0.999 & 0.880 & 0.388  \\
64  & \textbf{0.082} & 0.543 & 1.219 & 1.120 & 0.174  \\
70  & \textbf{0.040} & 0.473 & 0.837 & 1.010 & 0.090  \\
102 & \textbf{0.116} & 0.582 & 1.145 & 1.079 & 0.413  \\
106 & \textbf{0.114} & 0.637 & 1.016 & 0.958 & 0.195  \\
123 & \textbf{0.041} & 0.479 & 1.009 & 0.828 & 0.092  \\
127 & \textbf{0.095} & 0.645 & 1.051 & 1.022 & 0.389   \\
\hline\hline
\end{tabular}
\end{table}

\subsection{Large-Scale NRS\textbf{\textit{f}}M on CMU Motion Capture}
\label{sec:mocap}
\begin{figure*}[!t]
    \centering
    \includegraphics[width=0.095\linewidth]{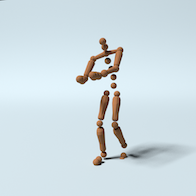}
    \includegraphics[width=0.095\linewidth]{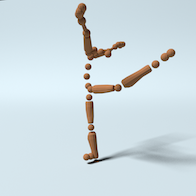}
    \includegraphics[width=0.095\linewidth]{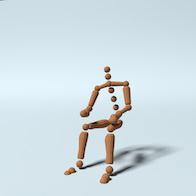}
    \includegraphics[width=0.095\linewidth]{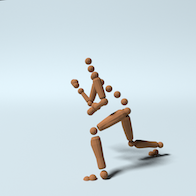}
    \includegraphics[width=0.095\linewidth]{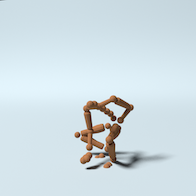}
    \includegraphics[width=0.095\linewidth]{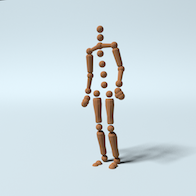}
    \includegraphics[width=0.095\linewidth]{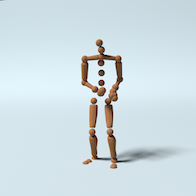}
    \includegraphics[width=0.095\linewidth]{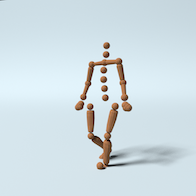}
    \includegraphics[width=0.095\linewidth]{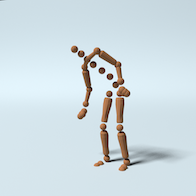}
    \includegraphics[width=0.095\linewidth]{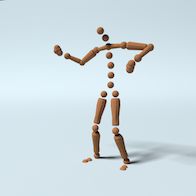}
    
    \vspace{1mm}
    
    \includegraphics[width=0.095\linewidth]{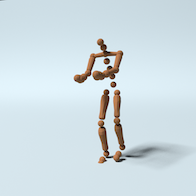}
    \includegraphics[width=0.095\linewidth]{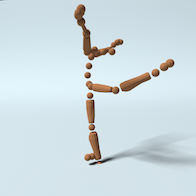}
    \includegraphics[width=0.095\linewidth]{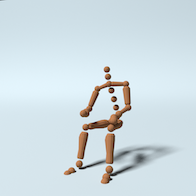}
    \includegraphics[width=0.095\linewidth]{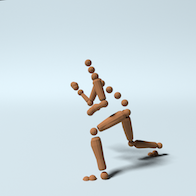}
    \includegraphics[width=0.095\linewidth]{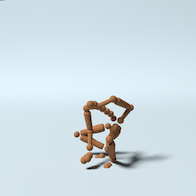}
    \includegraphics[width=0.095\linewidth]{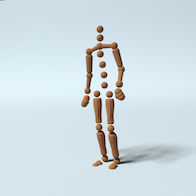}
    \includegraphics[width=0.095\linewidth]{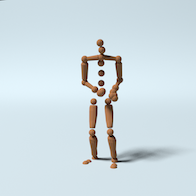}
    \includegraphics[width=0.095\linewidth]{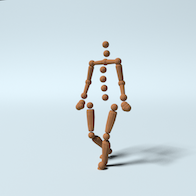}
    \includegraphics[width=0.095\linewidth]{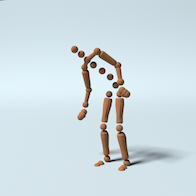}
    \includegraphics[width=0.095\linewidth]{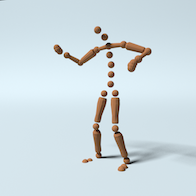}
    
    \vspace{1mm}
    
    \includegraphics[width=0.095\linewidth]{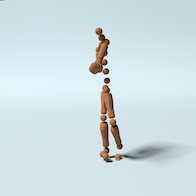}
    \includegraphics[width=0.095\linewidth]{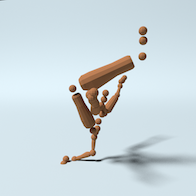}
    \includegraphics[width=0.095\linewidth]{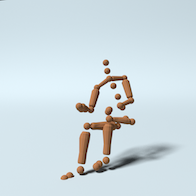}
    \includegraphics[width=0.095\linewidth]{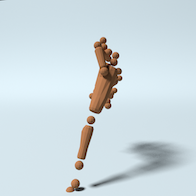}
    \includegraphics[width=0.095\linewidth]{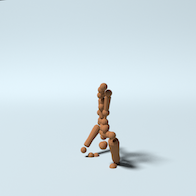}
    \includegraphics[width=0.095\linewidth]{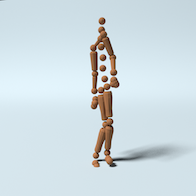}
    \includegraphics[width=0.095\linewidth]{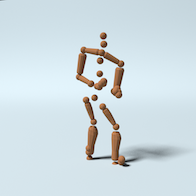}
    \includegraphics[width=0.095\linewidth]{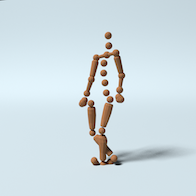}
    \includegraphics[width=0.095\linewidth]{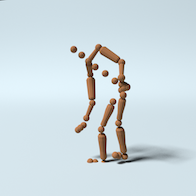}
    \includegraphics[width=0.095\linewidth]{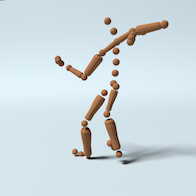}
    
    \vspace{1mm}
    
    \includegraphics[width=0.095\linewidth]{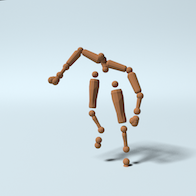}
    \includegraphics[width=0.095\linewidth]{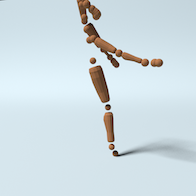}
    \includegraphics[width=0.095\linewidth]{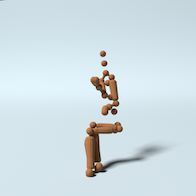}
    \includegraphics[width=0.095\linewidth]{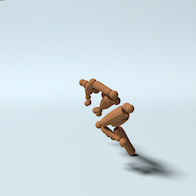}
    \includegraphics[width=0.095\linewidth]{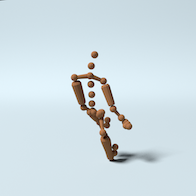}
    \includegraphics[width=0.095\linewidth]{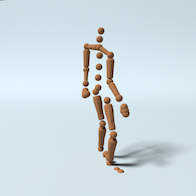}
    \includegraphics[width=0.095\linewidth]{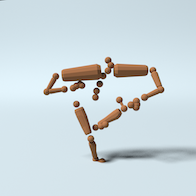}
    \includegraphics[width=0.095\linewidth]{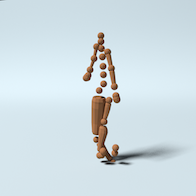}
    \includegraphics[width=0.095\linewidth]{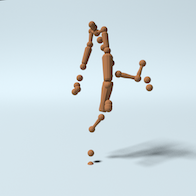}
    \includegraphics[width=0.095\linewidth]{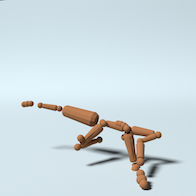}
    
    \vspace{1mm}
    
    \includegraphics[width=0.095\linewidth]{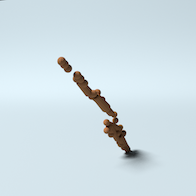}
    \includegraphics[width=0.095\linewidth]{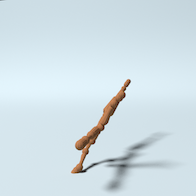}
    \includegraphics[width=0.095\linewidth]{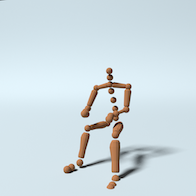}
    \includegraphics[width=0.095\linewidth]{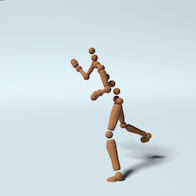}
    \includegraphics[width=0.095\linewidth]{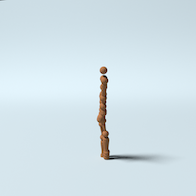}
    \includegraphics[width=0.095\linewidth]{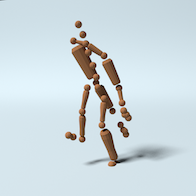}
    \includegraphics[width=0.095\linewidth]{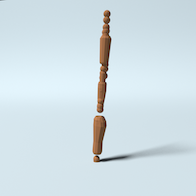}
    \includegraphics[width=0.095\linewidth]{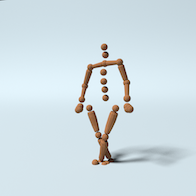}
    \includegraphics[width=0.095\linewidth]{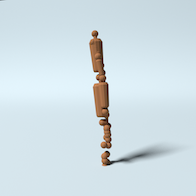}
    \includegraphics[width=0.095\linewidth]{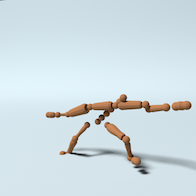}
    \caption{Qualitative evaluation on CMU Motion Capture dataset. From top to bottom are ground-truth, and
    respectively reconstructions by ours, CNS~\cite{lee2016consensus}, SPS~\cite{kong2016prior},
    NLO~\cite{del2007non}. From left to right are a randomly sampled frame from subjects 1, 5, 18, 23, 64, 70, 
    102, 106, 123, 127. In each rendering, spheres are reconstructed landmarks but bars are for descent 
    visualization. In each reconstruction, 3D shapes are alighted to the ground-truth by a orthonormal
    matrix. }
    \label{fig:mocap}
\end{figure*}

To evaluate the performance of our method on a large scale image sequence, we apply our method to 
solving the problem of \nrsfm{}, using the CMU motion capture dataset\footnote{http://mocap.cs.cmu.edu/}. 
We randomly select 10 subjects out of 144, and for each subject, we concatenate $80\%$ of motions to 
form large image collections and leave the remaining $20\%$ as unseen motions for testing generalization.
In this experiment, each subject contains more than ten thousand frames under randomly generated
orthogonal projections. We compare our method against state-of-the-art methods, summarized in Table~\ref{tab:mocap}. Due to the huge volume of frames, KSTA~\cite{gotardo2011kernel},
BMM~\cite{dai2014simple}, MUS~\cite{agudo2018image}, RIKS~\cite{hamsici2012learning}, and 
SFC~\cite{kong2016sfc} all fail and thus are omitted in the table. We also report the normalized mean 
3D error on unseen motions, labeled as UNSEEN. From Table~\ref{tab:mocap}, one can see that our method
obtains impressive reconstruction performance and outperforms all others again in every sequence. 
Moreover, our network generalizes well with unseen data, which implies the potential utility of our model to 
the application of single image 3D reconstruction. For qualitative evaluation, we randomly select a frame 
from each subject and render the reconstructed human skeleton in \figurename~\ref{fig:mocap}, which 
visually verifies the impressive performance of our deep solution.

\begin{figure}[!t]
    \centering
    \includegraphics[width=\linewidth]{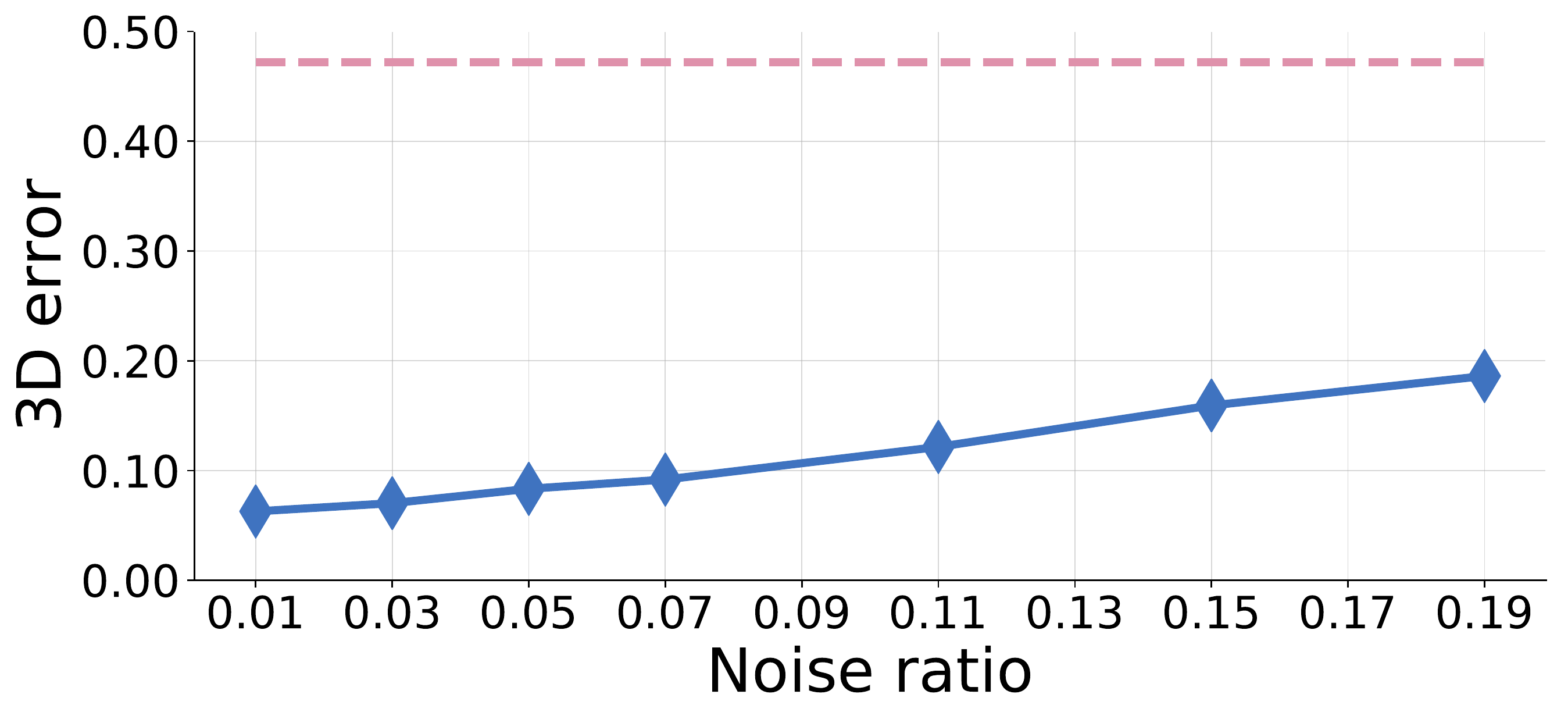}
    \caption{Normalized mean 3D error on CMU Motion Capture dataset with Gaussian noise perturbation. The
    blue solid line is ours while the red dashed line is CNS~\cite{lee2016consensus}, the lowest error
    of state-of-the-arts with \emph{no} noise perturbation.}
    \label{fig:noise}
\end{figure}

\subsubsection{Robustness analysis}
To analyze the robustness of our method, we retrain the neural network for Subject 70, using projected 
points perturbed by Gaussian noise. The results are summarized in \figurename~\ref{fig:noise}. The noise
ratio is defined as $\Vert \text{noise} \Vert_F / \Vert \Wv \Vert_F$. One can see that the error increases
slowly while adding a higher magnitude of noise; when adding up to $20\%$ noise to image coordinates, 
our method in blue still achieves better reconstruction compared to the best baseline with no noise
perturbation (in red). This experiment clearly demonstrates the robustness of our model and its high
accuracy against state-of-the-art works.

\subsubsection{Explicitly solve translation}
\begin{figure}[!t]
    \centering
    \includegraphics[width=\linewidth]{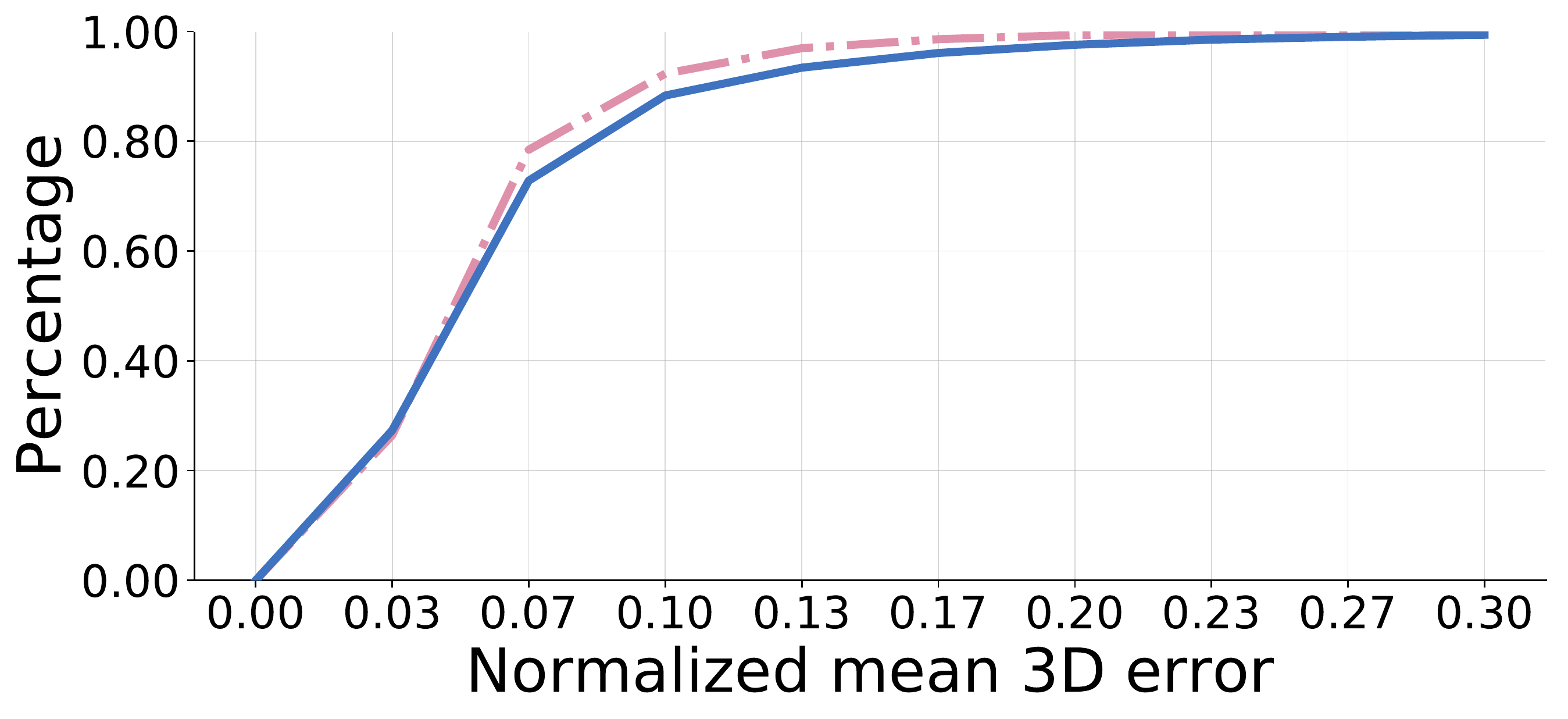}
    \caption{Percentage below a certain normalized mean 3D error. The blue solid line is our 4-by-2
    block sparse model, proposed to solve translation explicitly. The red dashed line is our 3-by-2
    block sparse model, applied on zero-centered data. These two plots are mostly identical.}
    \label{fig:translation}
\end{figure}

In this experiment, we verify the performance of the proposed 4-by-2 block sparse model. We focus on 
Subject 23, following the same experiment setting as above, except adding randomly generated translation.
To avoid removing translation, we do not normalize 2D correspondences. We then apply the proposed 4-by-2
block sparse model to the data with translation and compare it to the 3-by-2 block sparse model without 
translation. The normalized mean 3D error of the 4-by-2 model is 0.060, which is very close to the error without 
translation, \ie~0.054, and lower than state-of-the-arts without translation in the order of magnitude, 
as shown in Table~\ref{tab:mocap}.  To give a clearer sense of the quality of the reconstructed 3D shape, we 
draw two cumulative error plots in \figurename~\ref{fig:translation} that show the percentage of frames 
below a certain normalized mean 3D error. The two plots are mostly identical, implying the success 
of our 4-by-2 model.

\begin{figure}[!t]
    \centering
    \includegraphics[width=\linewidth]{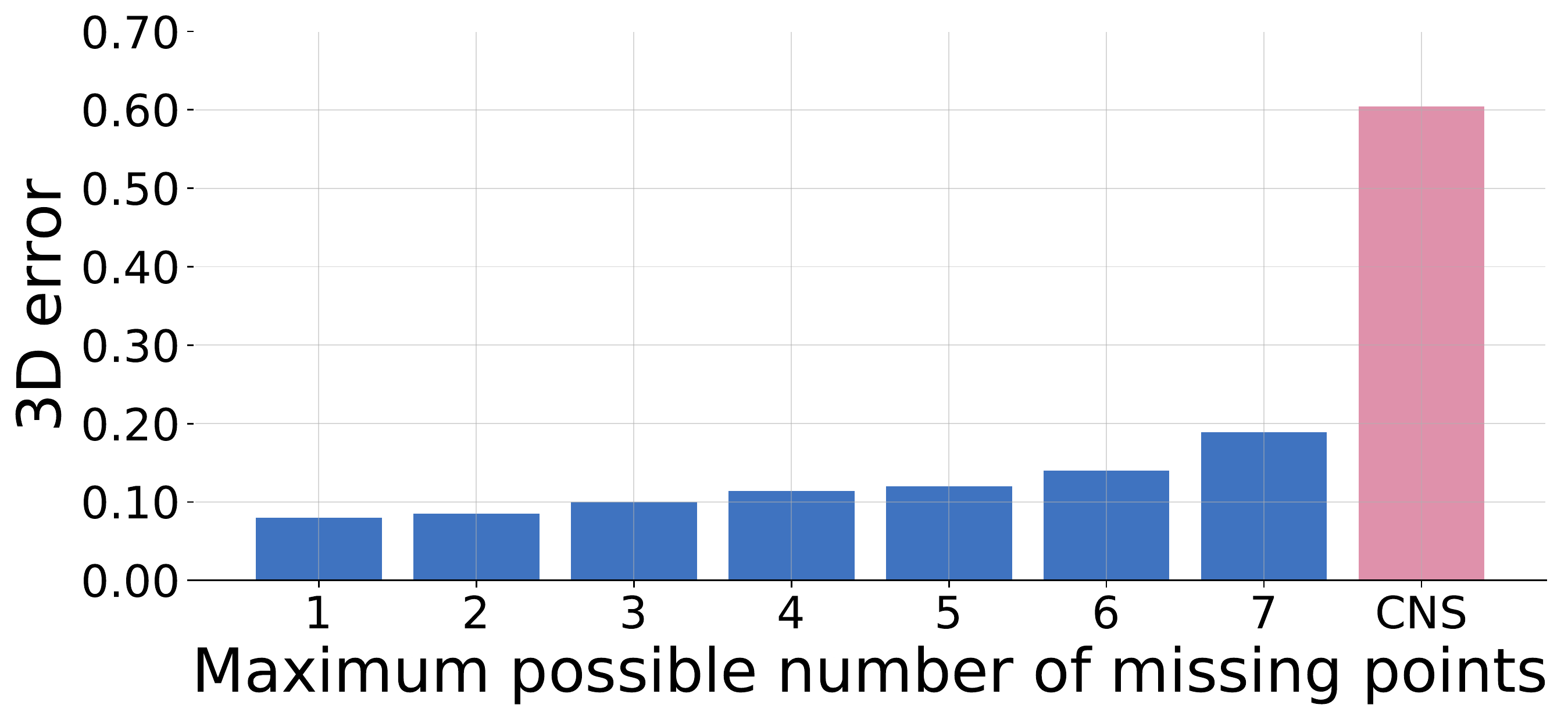}
    \caption{Normalized mean 3D error v.s. maximum possible number of missing points. Maximum possible
    number of missing points equals to three denotes every frame has to have one, two, or three missing
    points. The blue bar is our proposed network. The red bar is the best baseline when all points 
    are visible, \ie~CNS in Table~\ref{tab:mocap}.}
    \label{fig:missing}
\end{figure}

\subsubsection{Missing points}
In this experiment, we explore the capability of handling missing data. We focus on Subject 23 under 
orthogonal projection and 
sequentially train and test our proposed network on data with a different percentage of missing points.
Specifically, we control the maximum possible number of missing points and evaluate the performance from 
one to seven out of 31 total points. For example, when the maximum possible number of missing points is
three, then each frame has to have one, two, or three missing points in uniform distribution. We 
visualize the normalized mean 3D error in each case in  \figurename~\ref{fig:missing} and append the 
lowest error achieved by state-of-the-arts under the complete measurement assumption as a baseline. One can 
see that the 3D error increases when the maximum possible number of missing points grows. However, even 
making approximately 20\% (7/31) of points invisible, our proposed method still outperforms the best 
baseline with no missing points, \ie~CNS 0.604 in Table~\ref{tab:mocap}.

\subsubsection{Coherence as guide}
\begin{figure}[!t]
    \centering
    \includegraphics[width=\linewidth]{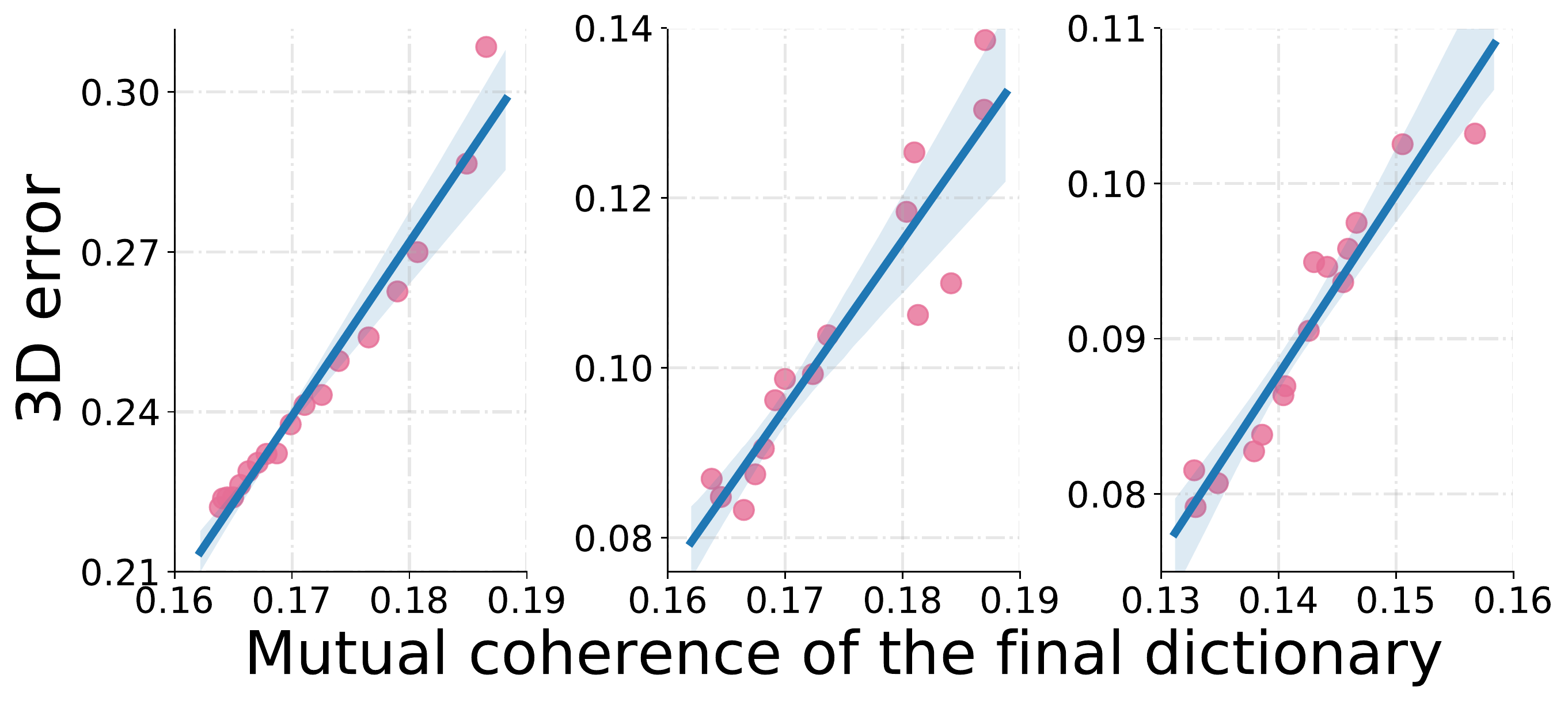}
    \caption{A scatter plot of the normalized mean 3D error v.s. the coherence of the final dictionary.
    The blue line is fitted based on the red points. Shading presents the quality of linear regression.
    From left to right are, respectively, for Subjects 5, 18, and 64.}
    \label{fig:coherence}
\end{figure}

Over-fitting is commonly observed in the deep learning community, especially in the \nrsfm{} area, where 
over-fitting to 2D correspondences will dramatically hurt the quality of reconstructions.
To solve this problem, we borrow a tool from compressed sensing -- mutual coherence~\cite{donoho2006stable}.
Mutual coherence measures the similarity between atoms in a dictionary. It is often used to depict 
the dictionary quality and build the bounds of sparse code reconstructability.
During training for each subject, we compute the normalized mean 3D error and the coherence of the last
dictionary in a fixed training iteration interval. By drawing the scatter plot of the error and the coherence, 
we observe a strong correlation, shown in \figurename~\ref{fig:coherence}. This implies that the
coherence of the final dictionary could be used as a measure of model quality.

Recall the proposed block sparse model in (\ref{eq:mlbsc_w}), wherein every block sparse code~$\Psiv_i$ is 
constrained by its subsequent representation and thus, the quality of code recovery depends not only on 
the quality of the corresponding dictionary but also the subsequent layers.  However, this is not 
applicable to the final code~$\Psiv_N$, making it overly reliant upon the final dictionary~$\Dv_N$.
From this perspective, the quality of the final dictionary measured by mutual coherence could serve as 
a lower bound of the entire system.  With the help of the coherence, we could avoid over-fitting even
when 3D evaluation is not available. This improves the utility of our deep \nrsfm{} in applications 
without 3D ground-truth.

\begin{figure*}[!t]
    \centering
    \includegraphics[width=0.12\linewidth]{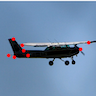}
    \includegraphics[width=0.12\linewidth]{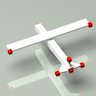}
    \includegraphics[width=0.12\linewidth]{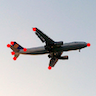}
    \includegraphics[width=0.12\linewidth]{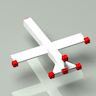}
    \includegraphics[width=0.12\linewidth]{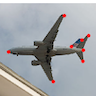}
    \includegraphics[width=0.12\linewidth]{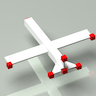}
    \includegraphics[width=0.12\linewidth]{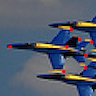}
    \includegraphics[width=0.12\linewidth]{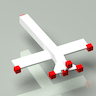}
    
    \vspace{1mm}
    
    \includegraphics[width=0.12\linewidth]{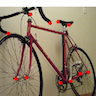}
    \includegraphics[width=0.12\linewidth]{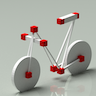}
    \includegraphics[width=0.12\linewidth]{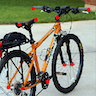}
    \includegraphics[width=0.12\linewidth]{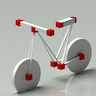}
    \includegraphics[width=0.12\linewidth]{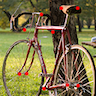}
    \includegraphics[width=0.12\linewidth]{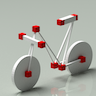}
    \includegraphics[width=0.12\linewidth]{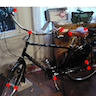}
    \includegraphics[width=0.12\linewidth]{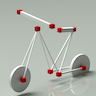}
    
    \vspace{1mm}
    
    \includegraphics[width=0.12\linewidth]{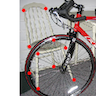}
    \includegraphics[width=0.12\linewidth]{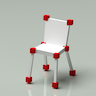}
    \includegraphics[width=0.12\linewidth]{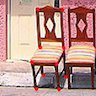}
    \includegraphics[width=0.12\linewidth]{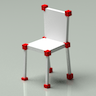}
    \includegraphics[width=0.12\linewidth]{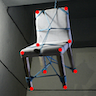}
    \includegraphics[width=0.12\linewidth]{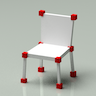}
    \includegraphics[width=0.12\linewidth]{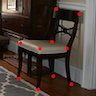}
    \includegraphics[width=0.12\linewidth]{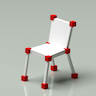}
    
    \vspace{1mm}
    
    \includegraphics[width=0.12\linewidth]{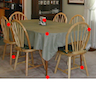}
    \includegraphics[width=0.12\linewidth]{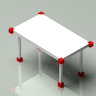}
    \includegraphics[width=0.12\linewidth]{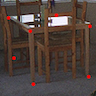}
    \includegraphics[width=0.12\linewidth]{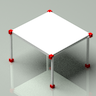}
    \includegraphics[width=0.12\linewidth]{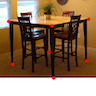}
    \includegraphics[width=0.12\linewidth]{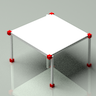}
    \includegraphics[width=0.12\linewidth]{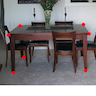}
    \includegraphics[width=0.12\linewidth]{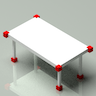}
    
    \caption{Qualitative evaluation on real images with hand-annotated 2D correspondences. Some images have
    missing points, due to occlusion. From top to bottom are aeroplanes, bicycles, chairs, and dining tables.
    For each pair, the left is an image with key points in red and the right is our reconstruction. In each
    rendering of reconstruction, red cubes are reconstructed points, but the planes and bars are manually 
    added for descent visualization. Our method successfully captures shape variations presented in the 
    images, \eg~table width-length ratio, the position of aeroplane wings, bicycle handlebar, and so forth.}
    \label{fig:real}
\end{figure*}

\subsection{Real Images}
Our proposed network is designed for applications on large-scale image sequences of highly deformable 
objects, especially object categories. However, to our best knowledge, commonly-used object datasets 
mostly contain less than one hundred images of reasonable quality, a number which is greatly insufficient 
to train a neural network. For example, most objects in the PASCAL3D dataset have more than 50\%
occluded points. To demonstrate the performance of our proposed network, we apply the model pre-trained
on synthetic images to real images with hand annotated correspondences. Due to the absence of 3D ground-truth,
we qualitatively evaluate the reconstructed shapes and show them in \figurename~\ref{fig:real}. One can see
that our proposed neural network successfully reconstructs the 3D shape for each image and impressively 
captures the subtle shape variation presented in the image, \eg~the table width-length ratio, the position of
aeroplane wings, the bicycle handlebar and so forth.

\section{Discussion}
This paper utilizes the weak-perspective projection to approximate the perspective projection.  
Though still not perspective, the jump from orthogonal projection (used in our previous
work~\cite{kong2019deep}) to weak-perspective allows us to apply our proposed neural network to practical
scenarios, \eg~real images with hand-annotated landmarks.  This feasibility is substantially a larger 
improvement than increasing accuracy via the perspective projection.  As for the future research, we believe,
rather than our used feed-forward network, a recurrent network could be more appropriate to handle the 
non-linearity of the perspective projection.  Moreover, a recurrent network could also provides a better
chance to handle a large portion of missing points. We believe, it is also promising to apply the basic 
ideas of this paper to the dual space, trajectory reconstruction. Due to the continuity of object trajectory,
introducing convolution to the hierarchical sparse model is an interesting attempt.

\section{Conclusion}
In this paper, we proposed to use the hierarchical sparse coding as a novel prior assumption for 
representing 3D non-rigid shapes and designed an innovative encoder-decoder neural network to solve 
the problem of \nrsfm{}. The proposed deep neural network was derived by generalizing the classical 
sparse coding algorithm, ISTA, to a block sparse scenario. The proposed network architecture
is mathematically interpretable as a hierarchical block sparse dictionary learning solver. Extensive
experiments demonstrated our superior performance against the state-of-the-art methods on various 
configurations including orthogonal projections, weak perspective projection, noise perturbations, 
missing points, real images, and even unseen shape variations. Finally, we propose to use the coherence 
of the final dictionary as a generalization measure, offering a practical way to avoid over-fitting and
select the best model without 3D ground-truth.

\ifCLASSOPTIONcompsoc
  \section*{Acknowledgments}
\else
  \section*{Acknowledgment}
\fi

This material is based upon work supported by the National Science Foundation under Grant No.1526033.

\ifCLASSOPTIONcaptionsoff
  \newpage
\fi



%
\bibliography{bib.bib}{}
\bibliographystyle{IEEEtran}

%

\begin{IEEEbiography}[{\includegraphics[width=1in,height=1.25in,clip,keepaspectratio]{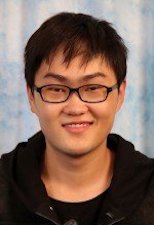}}]{Chen Kong}
is currently working toward the PhD degree at the Robotics Institute at Carnegie Mellon University. He 
received his Bechelor's degree from Tsinghua University, Beijing, China, in 2014. His research interests
include computer vision and machine learning, especially non-rigid structure from motion and model-based
object 3D reconstruction. He is a student member of the IEEE.
\end{IEEEbiography}

\begin{IEEEbiography}[{\includegraphics[width=1in,height=1.25in,clip,keepaspectratio]{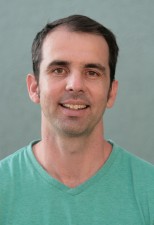}}]{Simon Lucey}
received the PhD degree from the Queensland University of Technology, Brisbane, Australia, in 2003. He is an Associate Research Professor at the Robotics Institute at Carnegie Mellon University. He also holds an adjunct professorial position at the Queensland University of Technology. His research interests include computer vision and machine learning, and their application to human behavior. He is a member of the IEEE.
\end{IEEEbiography}



\vfill


\end{document}